\documentclass[lettersize,journal]{IEEEtran}
\usepackage{amsmath,amsfonts}
\usepackage{algorithmic}
\usepackage{algorithm}
\usepackage{array}
\usepackage[caption=false,font=footnotesize,labelfont=rm,textfont=rm]{subfig}
\usepackage{textcomp}
\usepackage{stfloats}
\usepackage{url}
\usepackage{verbatim}
\usepackage{graphicx}
\usepackage{cite}
\usepackage{orcidlink}
\usepackage{tikz}
\usepackage{tabularx}
\usepackage{multirow}
\usepackage{ulem}
\usepackage{titlesec}
\usepackage{indentfirst}
\hypersetup{
colorlinks=true,
linkcolor = red,
citecolor = green,
urlcolor  = blue,
}

\titlespacing*{\section}{0pt}{5pt}{5pt}
\titlespacing*{\subsection}{0pt}{5pt}{5pt}
\begin{document}

\title{DFVO: Learning Darkness-free Visible and Infrared Image Disentanglement and Fusion All at Once}

\author{ Qi Zhou, Yukai Shi, Xiaojun Yang\IEEEauthorrefmark{1}, Xiaoyu Xian, Lunjia Liao, Ruimao Zhang,  and Liang Lin,~\IEEEmembership{Fellow, IEEE} 
\thanks{Q. Zhou and Y. Shi are with the School of Information Engineering, Guangdong University of Technology, Guangzhou, 510006, China (e-mail: zhouqi0530@gmail.com, ykshi@gdut.edu.cn).}
\thanks{X. Yang is with the School of Information Engineering, Guangdong University of Technology, Guangzhou, 510006, China, and also with the Key Laboratory of Photonic Technology for Integrated Sensing and Communication, Ministry of Education of China, Guangzhou, 510006, China (e-mail: yangxj18@gdut.edu.cn).}
\thanks{X. Xian is with CRRC Institute Co., Ltd., Beijing, China (e-mail: xxy@crrc.tech).}
\thanks{L. Liao is with the UBTECH Robotics Co., Ltd, Shenzhen, 518071, China (e-mail: lunjia.liao@ubtrobot.com).}
\thanks{R. Zhang is with the School of Data Science, The Chinese University of Hong Kong, Shenzhen, 518000, China. (e-mail: zhangruimao@cuhk.edu.cn).}
\thanks{L. Lin is with School of Data and Computer Science, Sun Yat-sen University, Guangzhou, 510006, China (e-mail: linliang@ieee.org).}}



\maketitle

\begin{abstract}
Visible and infrared image fusion is one of the most crucial tasks in the field of image fusion, aiming to generate fused images with clear structural information and high-quality texture features for high-level vision tasks. However, when faced with severe illumination degradation in visible images, the fusion results of existing image fusion methods often exhibit blurry and dim visual effects, posing major challenges for autonomous driving. To this end, a Darkness-Free network is proposed to handle Visible and infrared image disentanglement and fusion all at Once (DFVO), which employs a cascaded multi-task approach to replace the traditional two-stage cascaded training (enhancement and fusion), addressing the issue of information entropy loss caused by hierarchical data transmission. Specifically, we construct a latent-common feature extractor (LCFE) to obtain latent features for the cascaded tasks strategy. Firstly, a details-extraction module (DEM) is devised to acquire high-frequency semantic information. Secondly, we design a hyper cross-attention module (HCAM) to extract low-frequency information and preserve texture features from source images. Finally, a relevant loss function is designed to guide the holistic network learning, thereby achieving better image fusion. Extensive experiments demonstrate that our proposed approach outperforms state-of-the-art alternatives in terms of qualitative and quantitative evaluations. Particularly, DFVO can generate clearer, more informative, and more evenly illuminated fusion results in the dark environments, achieving best performance on the LLVIP dataset with 63.258 dB PSNR and 0.724 CC, providing more effective information for high-level vision tasks. Our code is publicly accessible at \href{https://github.com/DaVin-Qi530/DFVO}{https://github.com/DaVin-Qi530/DFVO}.
\end{abstract}


\section{Introduction}
\label{sec:intro}
\IEEEPARstart{W}{ith} the advancement of imaging technologies, visible sensor has become the preferred choice for environmental perception in the field of autonomous driving. However, when faced with the problem of visible illumination degradation, single-modality sensor often fails to accurately provide a comprehensive description of the entire scene. Therefore, image fusion algorithms are emerging, which can extract the important information from paired visible-infrared images and merge it into fused images, thereby facilitating subsequent applications. Specifically, visible images contain meaningful color that conforms to human visual perception and rich texture information. By acquiring significant thermal radiation information, infrared images can maintain complete structural information even in harsh environments. Consequently, the primary task of visible and infrared image fusion is to generate fused images with rich texture details, distinct structural information, and better color perception, \textit{etc.}, which are preferably applied in high-level computer tasks such as object detection\cite{ma2021stdfusionnet,shi2024diff,lu2024sirst}, instance segmentation\cite{liu2023yolactfusion}, and autonomous navigation\cite{bhatnagar2015novel}, \textit{etc.}

Visible and infrared image fusion can be divided into two major categories based on fusion strategies: traditional approaches and deep learning-based approaches. The traditional methods can be further divided into five categories including multi-scale transform-\cite{da2006nonsubsampled,shi2025crossfuse}, saliency analysis-\cite{han2013fast}, subspace learning-\cite{fu2008multiple}, sparse coding-\cite{zhang2019infrared} and hybrid-based\cite{zang2021ufa} models. These approaches focus on how to use mathematical formulations to express and extract feature information from source images, and then use appropriate fusion strategies to merge the extracted feature information. Deep learning-based fusion strategies can be classified into convolutional neural  network (CNN)-based\cite{zhang2020ifcnn}, auto-encoder (AE)-based\cite{ren2021infrared}, generative adversarial network (GAN)-based\cite{li2023mrfddgan}, and Transformer-based\cite{li2022cgtf, wang2022swinfuse} methods. CNN-based and AE-based methods adhere to the rule of feature extraction, feature fusion, and feature reconstruction, but CNN-based methods emphasise the design of the loss function,  while AE-based methods focus on the feature extraction and reconstruction capabilities of auto-encoders. Transformer-based methods place more emphasis on the contextual expression ability of multi-modal feature in integration processes. GAN-based methods employ adversarial learning to generate fusion images that approximate the distribution of multi-modal information under unsupervised learning.

\begin{figure*}[h]
\centering
\includegraphics[width=0.96\linewidth]{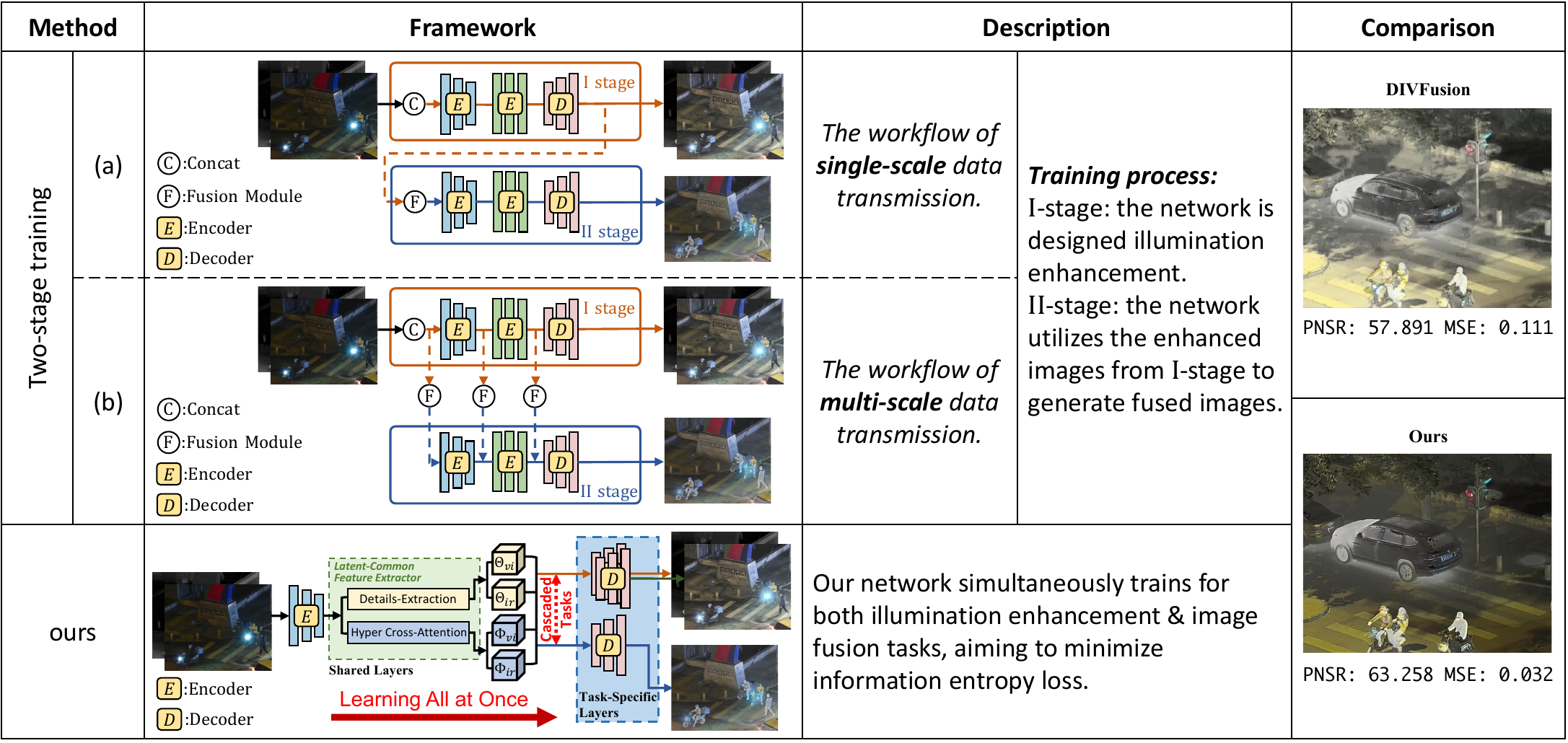}
\caption{Existing fusion methods in low-light environments \textit{vs.} DFVO. (a) The workflow of single-scale data transmission (\textit{such as} DIVFusion\cite{tang2023divfusion}, Ev-fusion\cite{zhang2024ev}, L2fusion\cite{gao2023l2fusion}, and LENFusion\cite{chen2024lenfusion}). (b) The workflow of multi-scale data transmission (\textit{such as} EFMN\cite{wang2023enlighten}). Among them, the compared images are the results of DIVFusion (upper) and our method (lower). \textcolor{blue}{"E"} denotes down-sampling, and \textcolor{green}{"E"} represents feature-expansion. }
\vspace{-15pt}
\label{fig:introduction}
\end{figure*}

Although deep learning-based methods can produce fused images with the significant effect, researchers prioritize the quality of fused images over other factors. However, when confronted with degradation caused by low-light conditions and corresponding pixel regions with missing information in infrared images, the fusion effect often becomes less satisfactory. To address this issue, darkness-free image fusion methods \cite{zhang2024ev, gao2023l2fusion, chen2024lenfusion, wang2023enlighten,xian2024crose,shi2024nitedr,yang2023reference} are proposed. These approaches all utilize a two-stage cascaded training strategy to obtain fused images with good visual perception. More specifically, the two-stage training method represents the learning of two sub-module networks. The specific phases are as follows: 
\textit{\textbf{In the first stage:} the network is used for illumination enhancement, with the aim of disentangling scene illumination.}
\textit{\textbf{In the second stage:} the network is centered around image fusion using enhanced images, aiming to generate fused images suitable for low-light environments.}

As depicted in Fig. \ref{fig:introduction}(a), DIVFusion used single-scale data for information transmission between stages, leading to excessive-noise and over-saturation in fused results. In response, EFMN\cite{wang2023enlighten} proposed using multi-scale data for information transmission between stages, as illustrated in Fig. \ref{fig:introduction}(b). In Information Entropy\cite{carter2007introduction} theory, these cascaded-stage optimization methods lead to the gradual loss of input information at each stage and amplify the noise generated during the processing, thereby severely affecting the quality of the fused image, as seen in Fig. \ref{fig:introduction}. To address this, our method proposes an innovative holistic network that transforms the previous two-stage cascaded learning into a cascaded multi-task approach. By guiding the network through task-specific loss functions, DFVO aims to address scene illumination degradation and generate fused images with clear and abundant texture. The specific contributions can be summarized as follows:

\begin{itemize}
\item{We propose a learns-all-at-once network to simultaneously accomplish several critical tasks. We utilize a cascaded multi-task strategy to guide the network, enabling holistic learning of illumination enhancement \& image fusion in low-light environments.}
\item{A details-extraction module (DEM) is constructed to extract high-frequency semantic information, and a hyper cross-attention module (HCAM) is designed to extract fundamental feature information.}
\item{The task-specific loss functions are designed to enhance the visual quality of fused images, which helps reduce the impact of local over-exposure on the fused image.}
\item{The experiments show that our results exhibit greater clarity, balanced brightness, and color representation in low-light environments. }
\end{itemize}


\section{Related Works}
\label{sec:rw}


\subsection{DL-based Image Fusion}
In the domain of image fusion, CNN-, AE-, and GAN-based models all generate satisfactory fused results. DenseFuse\cite{li2018densefuse} and U2Fusion\cite{xu2020u2fusion} stood out as classic examples of CNN-based methods, which utilized residual networks and well-designed loss functions to preserve more deep features in the encoder. However, the relatively simple network architectures constrain their performance potential. GANMcC\cite{ma2020ganmcc} approached image fusion as a multi-distribution simultaneous estimation problem. Nevertheless, the inherent instability of GAN training remains a significant limitation. Recently, the integration of image fusion with high-level computational tasks has become a trend. For instance, TarDAL\cite{liu2022target} and SeAFusion\cite{tang2022image} leveraged detection and segmentation loss functions to guide the generation of fused images, making the fused images more suitable for autonomous driving applications. 

\subsection{Invertible Neural Networks}

The invertible neural networks (INNs) were initially proposed by NICE\cite{dinh2014nice} and have been effectively integrated into image processing fields such as image steganography\cite{li2023iscmis}, image compression\cite{xie2021enhanced}, image rescaling\cite{xiao2020invertible}, and image coloring\cite{ardizzone2019guided} due to their lossless information-preserving property.
In previous classification tasks\cite{zhuang2019invertible}, INNs have demonstrated the ability to reduce network memory while enhancing feature extraction capabilities. Therefore, we aim to leverage the ability of INNs to preserve lossless information for extracting high-frequency information from source images, enabling a more comprehensive feature extraction approach.

\subsection{Self-Attention Mechanisms}

Self-attention was first utilized in natural language processing (NLP)\cite{vaswani2017attention} tasks, where its fundamental idea is to selectively extract essential information from a large amount of feature and focus on these key elements while suppressing irrelevant parts, so as to improve the performance and accuracy of computer vision tasks\cite{li2020multigrained, chen2021defect}. In 2021, a multi-head self-attention model known as Vision Transformer (ViT)\cite{dosovitskiy2020image} was proposed, which focuses the network's attention on important image pixels through an attention mechanism, thus improving the network's sensitivity to key pixels, allowing it to have a stronger generalization ability than CNNs and to be more widely used in other computer vision tasks\cite{zhang2024perceive, zhang2022transformer, liu2023transcending, tian2023vision}. In the Section \ref{sec:latent-common feature extractor}, we incorporate improved attention modules to enhance the performance of feature extraction.

\section{Proposed Method}
\label{sec:method}


\subsection{Learning Cascaded Tasks All at Once}
\label{sec:learning cascaded tasks all at once}
\begin{figure*}[!h]
\centering
\includegraphics[width=0.9\linewidth]{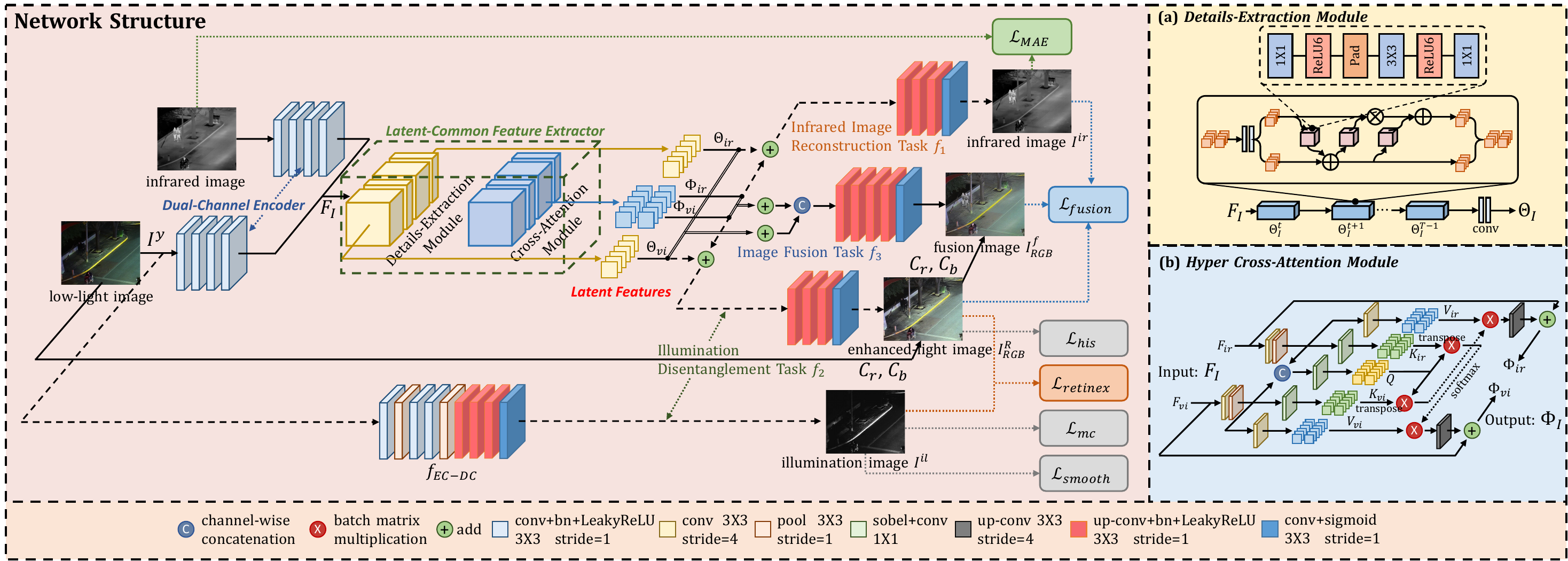}
\caption{The overall architecture of our method. The parallel cascaded tasks include the infrared image-reconstruction task, illumination disentanglement task, and image fusion task. (a) The specific structure of the Details-Extraction Module, which aims to capture high-frequency features from the source images. (b) The architecture of the Hyper Cross-Attention Module, which is designed to obtain the low-frequency features. }
\vspace{-15pt}
\label{fig:framework}
\end{figure*}

Our method adopts a one-stage cascaded multi-task machine, enabling the network to simultaneously accomplish several essential tasks. The overall working architecture is shown in Fig. \ref{fig:framework}, and the definitions of cascaded tasks $f_t$ are given as follows:

\textit{Infrared Image-Reconstruction Task:} Given that the shared layers should fully learn the feature information of the source images, and this task should facilitate the cascaded learning of the image fusion task. Additionally, the infrared images need to be reconstructed, and the entire task can be expressed as follows:
\begin{equation}
I^{ir} = f_{1}(\Theta_{ir} + \Phi_{ir}).
\end{equation}
Here, $I^{ir} \in \mathbb{R}^{H \times W \times 1}$ indicates the reconstructed infrared image. $f_{1}$ denotes the task-specific network for this task, which comprises four $3 \times 3$ convolutional layers, with the first three followed by transpose convolutional operations, batch normalization, and LeakyReLU activation function. The last layer is composed of a convolutional operation and a Sigmoid function. $\Theta_{ir}$ and $\Phi_{ir}$ are the outputs from the shared layers, representing latent infrared features. Through this task, the shared layers can retain richer infrared features, allowing them to provide better infrared images for the main image fusion task.

\textit{Illumination Disentanglement Task:}
Considering that generated images should not be obscured by darkness environments, this task employs the Retinex theory \cite{land1971lightness} at the feature level to generate the degraded illumination image, which can separate the illumination information of low-light images. Meanwhile, the enhanced image produced by this task can further contribute to the brightness enhancement of the fused image. The specific task can be defined as: 
\begin{equation}
\begin{aligned}
I^R &= f_{2}(\Theta_{vi} + \Phi_{vi}), \\
I^{il} &= f_{EC-DC}(I^y).
\end{aligned}
\end{equation}
Here, $I^R, I^{il} \in \mathbb{R}^{H \times W \times 1}$ represent the enhanced image and the illumination image, respectively. $I^y$ denotes the Y-channel of the visible image. $f_2$ signifies the task-specific network for the illumination disentanglement task, which has the same structure as $f_1$. $\Theta_{vi}$ and $\Phi_{vi}$ indicate latent visible features, which come from the shared layers. It is worth mentioning that the illumination image only learns the brightness information of the source images. Therefore, we design a separately configured encoder-decoder network $f_{EC-DC}$ to reduce the burden on this task. The specific structure of $f_{EC-DC}$ can be seen in Fig. \ref{fig:framework}. Through this task, we can obtain the enhanced image with suitable brightness and serve it to cascaded image fusion task.

 \textit{Image Fusion Task:}
As the primary task, we compel the holistic network to focus on fusing the extracted feature information. Specifically, we utilize latent features from the shared layers as fusion features, which facilitate the reduction of information entropy loss during data transmission. The image fusion task is introduced as follows:
\begin{equation}
I^f = f_{3}((\Theta_{ir} + \Theta_{vi}) \odot (\Phi_{ir} + \Phi_{vi})).
\end{equation}
Here, $I^f \in \mathbb{R}^{H \times W \times 1}$ signifies the fused image. $f_3$ represents the task-specific network for the image fusion task, which has a structure similar to $f_1$ but with an additional transpose convolutional layer. $\odot$ denotes the channel-wise concatenation operation. 

\subsection{Latent-Common Feature Extractor}
\label{sec:latent-common feature extractor}

\begin{figure}[!t]
\includegraphics[width=\linewidth]{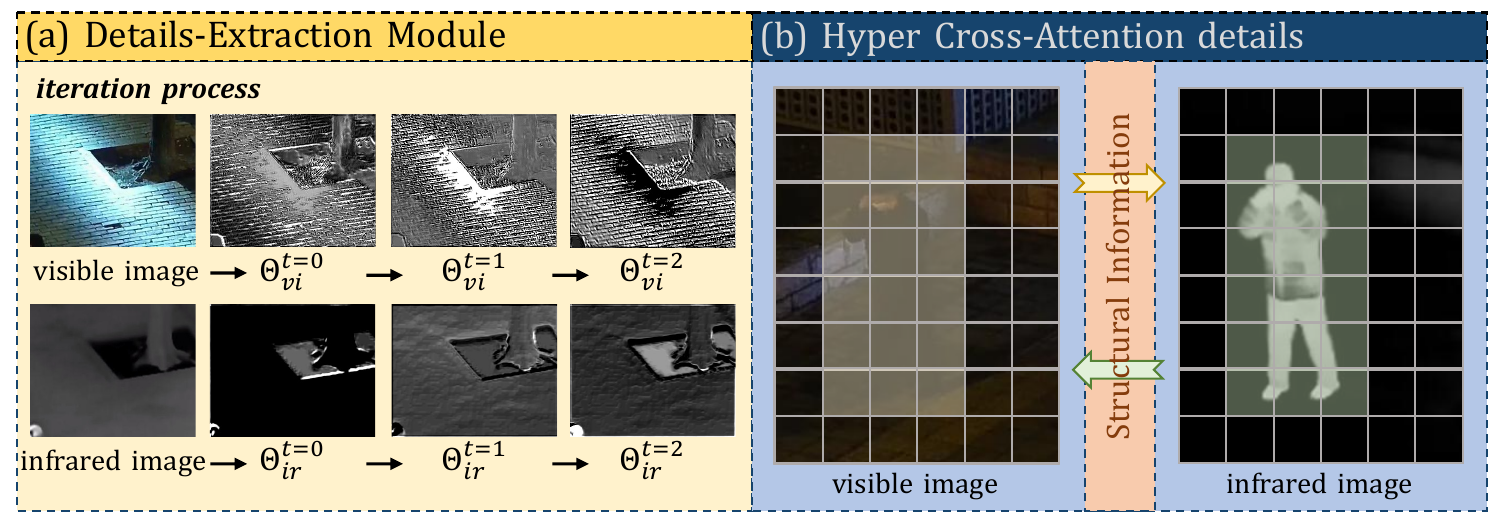}
\caption{(a) The visual results of iteration process in the Details-Extraction Module. (b)The interaction details of the Hyper Cross-Attention Module. }
\vspace{-15pt}
\label{fig:modules}
\end{figure}

\emph{1) Details-Extraction Module (DEM):} To better extract the semantic information distribution of the high-frequency detail features, a DEM is constructed. The network architecture of DEM is depicted in Fig. \ref{fig:framework}(a), and the iterative process it employs can be defined as:
\begin{equation}
\label{equ: inn block}
\resizebox{\linewidth}{!}{
$\begin{aligned}
\sum\limits_{n=1}^{C/2}{\Theta_{I}^{t+1}} &= \sum\limits_{n=1}^{C/2}{\Theta_{I}^{t}} + f_{BRB}(\sum\limits_{n=C/2}^{C}{\Theta_{I}^{t}}), \\
\sum\limits_{n=C/2}^{C}{\Theta_{I}^{t+1}} &=
f_{BRB}(\sum\limits_{n=1}^{C/2}{\Theta_{I}^{t+1}}) + \sum\limits_{n=C/2}^{C}{\Theta_{I}^{t}} \otimes e^{f_{BRB}(\sum\limits_{n=1}^{C/2}{\Theta_{I}^{t+1}})}.
\end{aligned}$}
\end{equation}
Here, $\otimes$ denotes the channel-wise multiplication operation, $C$ indicates the total number of channels in the features, $t$ represents the number of iterations per stage. $f_{BRB}$ signifies the bottleneck residual block (BRB)~\cite{sandler2018mobilenetv2}. Notably, the batch normalization (BN) layer within the BRB may impose abnormal constraints on the DEM, leading to the output of the DEM tending towards $0$. To address this issue, we isolate the BN layer from the BRB in the DEM. 

Given the features $\sum\limits_{n=1}^{C/2}{\Theta_{I}}$ and $\sum\limits_{n=C/2}^{C}{\Theta_{I}}$ obtained from Eq. (\ref{equ: inn block}), the detail features can be acquired as follows:
\begin{equation}
\begin{cases}
\Theta_{I}^{t} = A_t(F_{I}), & t = 0, \\
\Theta_{I}^{t+1} = A_{t+1}(\sum\limits_{n=1}^{C/2}{\Theta_{I}^{t+1}} \odot \sum\limits_{n=C/2}^{C}{\Theta_{I}^{t+1}}), & 0 < t < T, \\
\end{cases}
\end{equation}
where $A(\cdot)$ signifies the operation of removing speckle noise, which consists of a $3 \times 3$ convolutional layer. $F_{I}$ refers to the visible and infrared features obtained from the dual-channel encoder. $\odot$ indicates the concatenation in channel dimensions. $T=3$ represents the total number of iteration stages, and $\Theta_{I}^{t}$ is the output features of each iteration process. To observe the changes throughout the entire iteration process more intuitively, as shown in Fig. \ref{fig:modules}(a), the detail visible features $\Theta_{vi}$ and detail infrared features $\Theta_{ir}$ are gradually extracted, assisting cascaded tasks to generate images with more abundantly semantic information.

\emph{2) Hyper Cross-Attention Module (HCAM):} To aid DEM, the HCAM is designed to extract low-frequency features from source images, providing foundational information for cascaded tasks. The specific structure of HCAM is illustrated in Fig. \ref{fig:framework}(b), and the process can be expressed as follows:
\begin{equation}
\Phi_{vi} = f_{C}^{-1}(f_{CA_{vi}}) + F_{vi},
\label{equ: cam_vi}
\end{equation}
where $f_{C}^{-1}$ represents several transpose convolutional operations. $F_{vi}$ indicates the visible features from the dual-channel encoder. $f_{CA_{vi}}$ can guide the network to focus on specific regions, as defined by:
\begin{equation}
f_{CA_{vi}} = softmax(\frac{Q(|\nabla (\varphi_{vi} \odot \varphi_{ir})|) K(|\nabla \varphi_{ir}|)^T}{\sqrt{d_k}}) V(\varphi_{vi}).
\label{equ:cross-attention}
\end{equation}
Here, $d_k$ denotes the dimensionality of matrices. $|\cdot|$ represents the absolute value operation. $Q$, $K$, and $V$ indicate the query, key and value matrix, respectively. $\nabla$ signifies the Sobel operator. $\odot$ indicates the concatenation, which is employed to increase the information content of the query matrix $Q$. $\varphi_{vi}$ and $\varphi_{ir}$ are the inputted features of the $f_{CA_{vi}}$, which are computed by convolutional operation and a max pooling function. As shown in Fig. \ref{fig:modules}(b), the person is challenging to discern in the visible image, whereas the corresponding infrared image reveals clear structural information. Therefore, we employ the hyper cross-attention strategy to alleviate this issue. Concretely, we use infrared structural features $|\nabla \varphi_{ir}|$ as input to the key matrix $K$ in Eq. (\ref{equ:cross-attention}), allowing the HCAM to focus on multi-modal feature information in a hyper cross-attention manner. Moreover, the expression of $\Phi_{ir}$ is similar to $\Phi_{vi}$, which can be defined as:
\begin{equation}
\Phi_{ir} = f_{C}^{-1}(f_{CA_{ir}}) + F_{ir},
\label{equ: cam_ir}
\end{equation}
where $F_{ir}$ represents the infrared features from the dual-channel encoder. $f_{CA_{ir}}$ can be expressed as follows:
\begin{equation}
f_{CA_{ir}} = softmax(\frac{Q(|\nabla (\varphi_{vi} \odot \varphi_{ir})|) K(|\nabla \varphi_{vi}|)^T}{\sqrt{d_k}}) V(\varphi_{ir}).
\end{equation}
Eventually, through the calculation of HCAM, we can obtain a large amount of foundational feature information.

\subsection{Loss Function}
To ensure that the network parameters of our method are correctly updated, we design the loss function based on the cascaded tasks. The total loss function is as follows:
\begin{equation}
\mathcal{L}_{\theta}(w_{0}, w_{1}, \cdots, w_{t}) = \sum\limits_{t=1}^{T}{ \zeta_t(\mathcal{L}_t(w_0, w_t)). }
\label{equ: total loss}
\end{equation}
Here, $w_{0}$ represents the weights of the shared layers. $w_{t}$ are the task-specific weights. $t$ denotes the $t$-th task according to Section \ref{sec:learning cascaded tasks all at once}. $T$ = 3 indicates the total number of cascaded tasks. $\zeta_t$ signify the hyper-parameters that determine the importance of tasks. $\mathcal{L}_t(\cdot)$ represent the task-specific loss functions.

\begin{figure*}[!t]
\centering
\subfloat[Visible Image]{
		\includegraphics[width=0.96in]{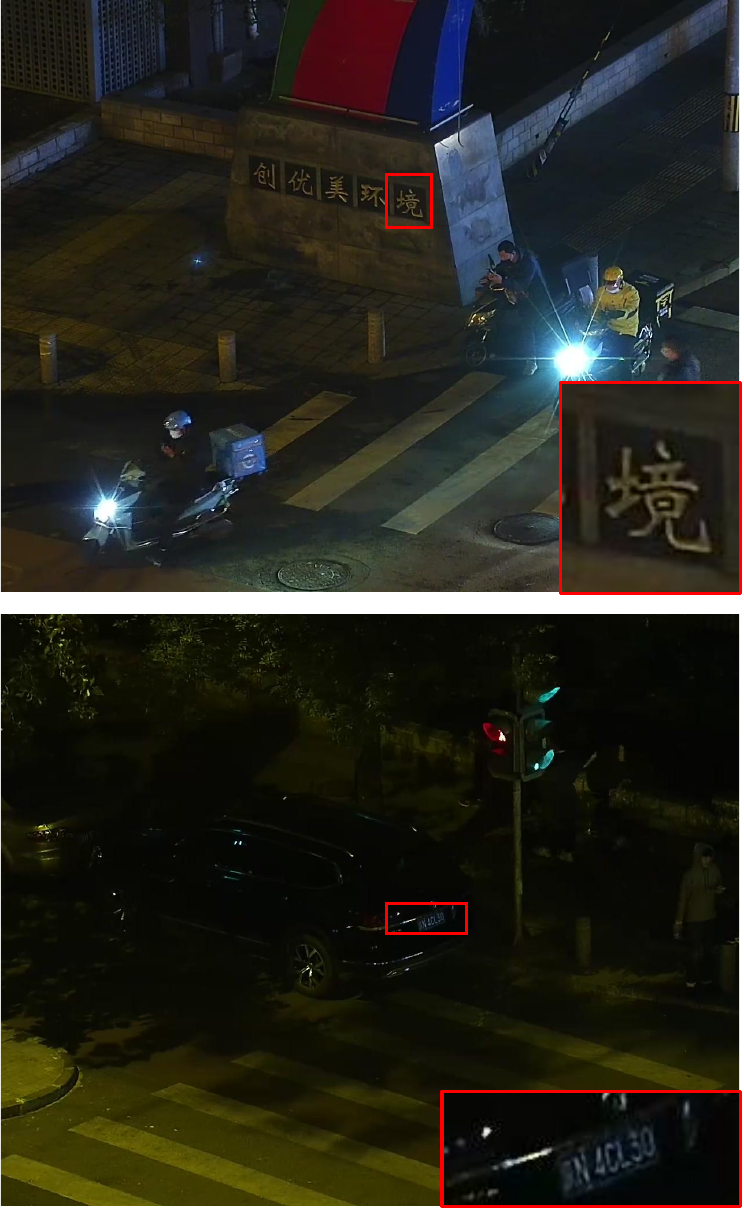}}
\subfloat[Infrared Image]{
		\includegraphics[width=0.96in]{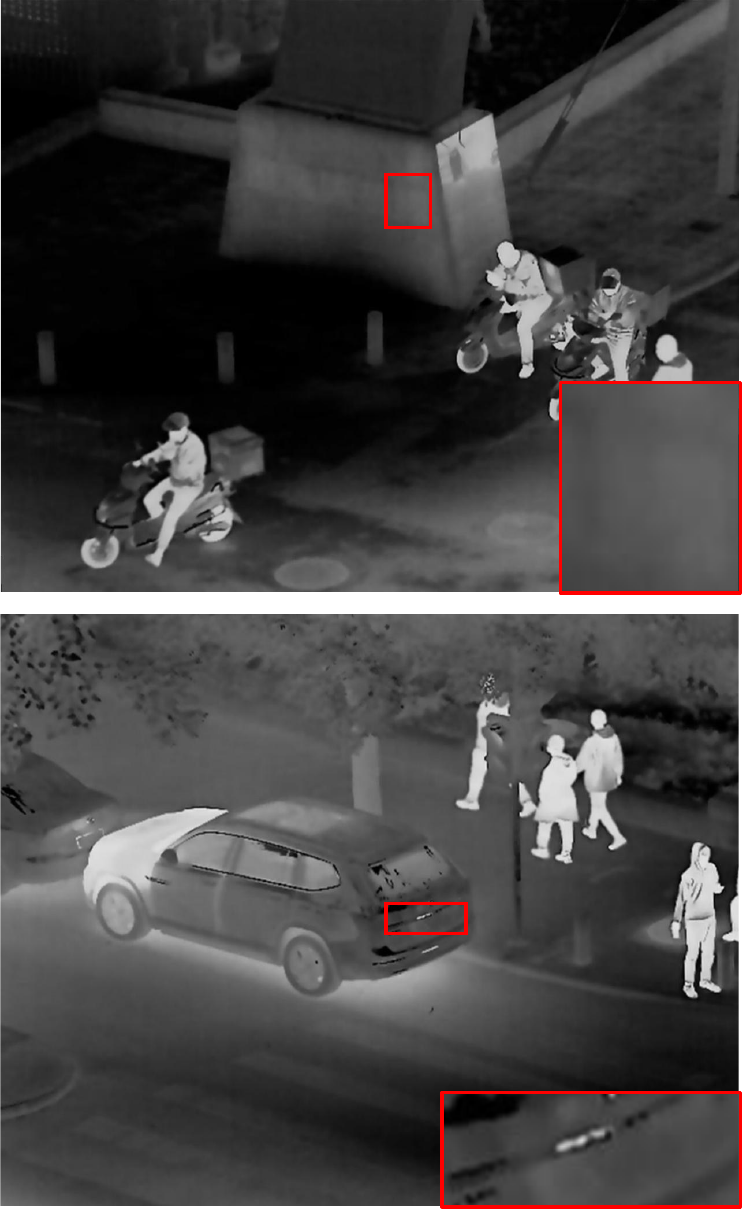}}
\subfloat[SeAFusion]{
		\includegraphics[width=0.96in]{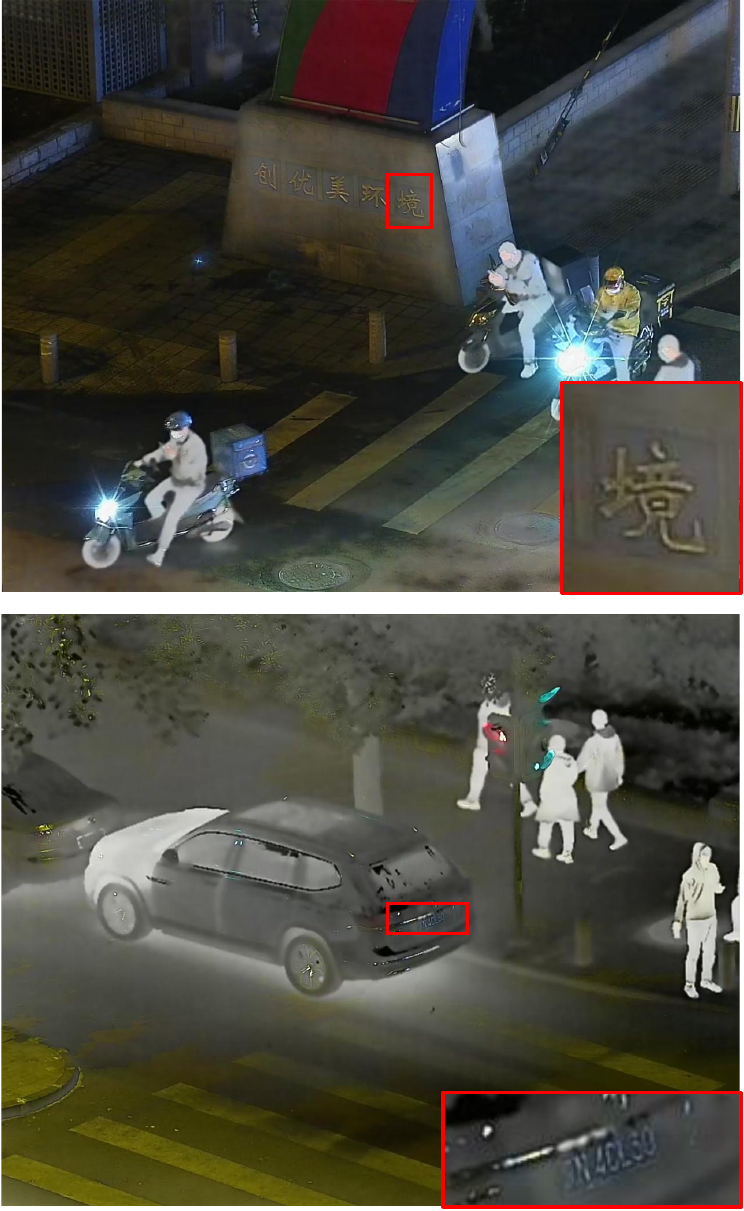}}
\subfloat[U2Fusion]{
		\includegraphics[width=0.96in]{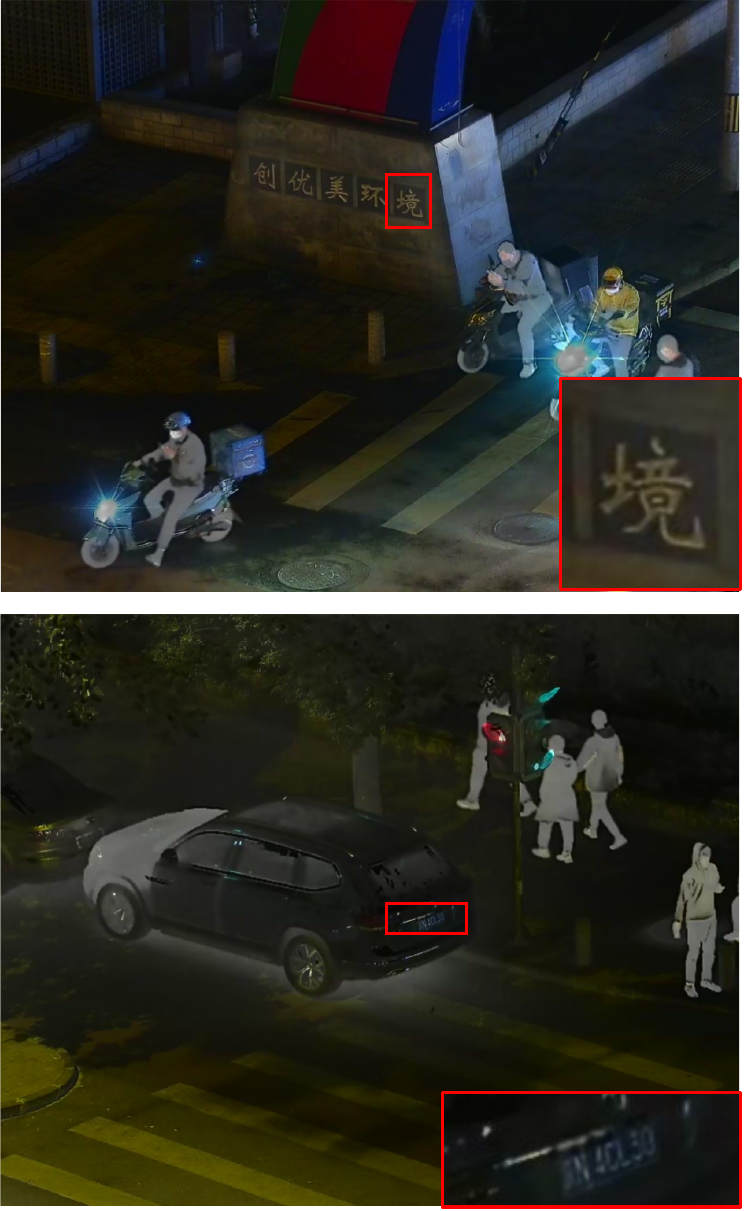}}
\subfloat[DenseFuse]{
		\includegraphics[width=0.96in]{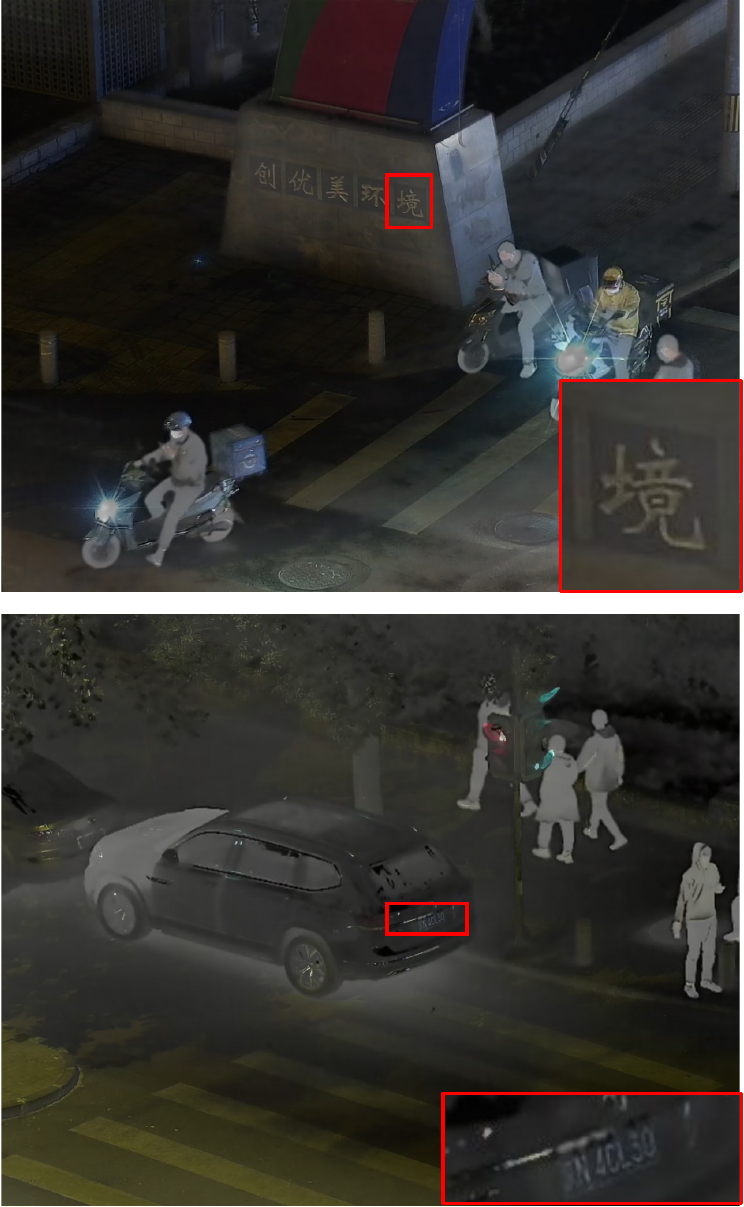}}
\subfloat[TarDAL]{
		\includegraphics[width=0.96in]{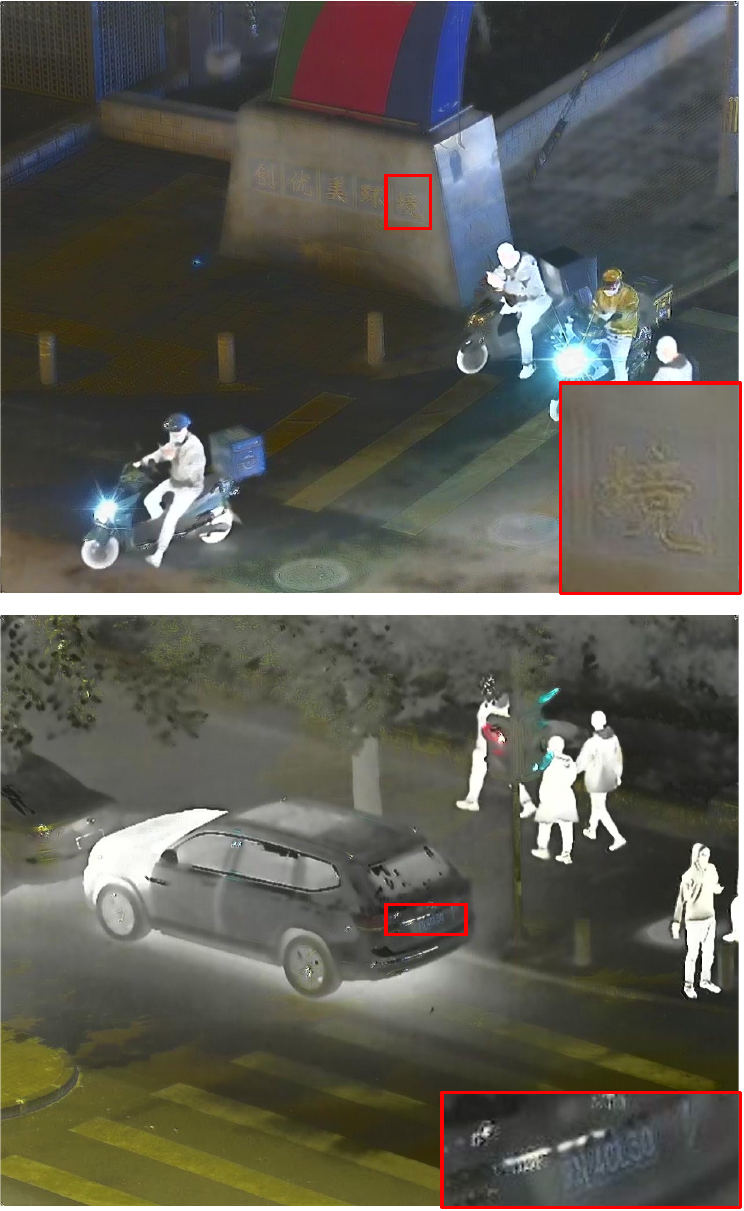}}
\subfloat[Ours]{
		\includegraphics[width=0.96in]{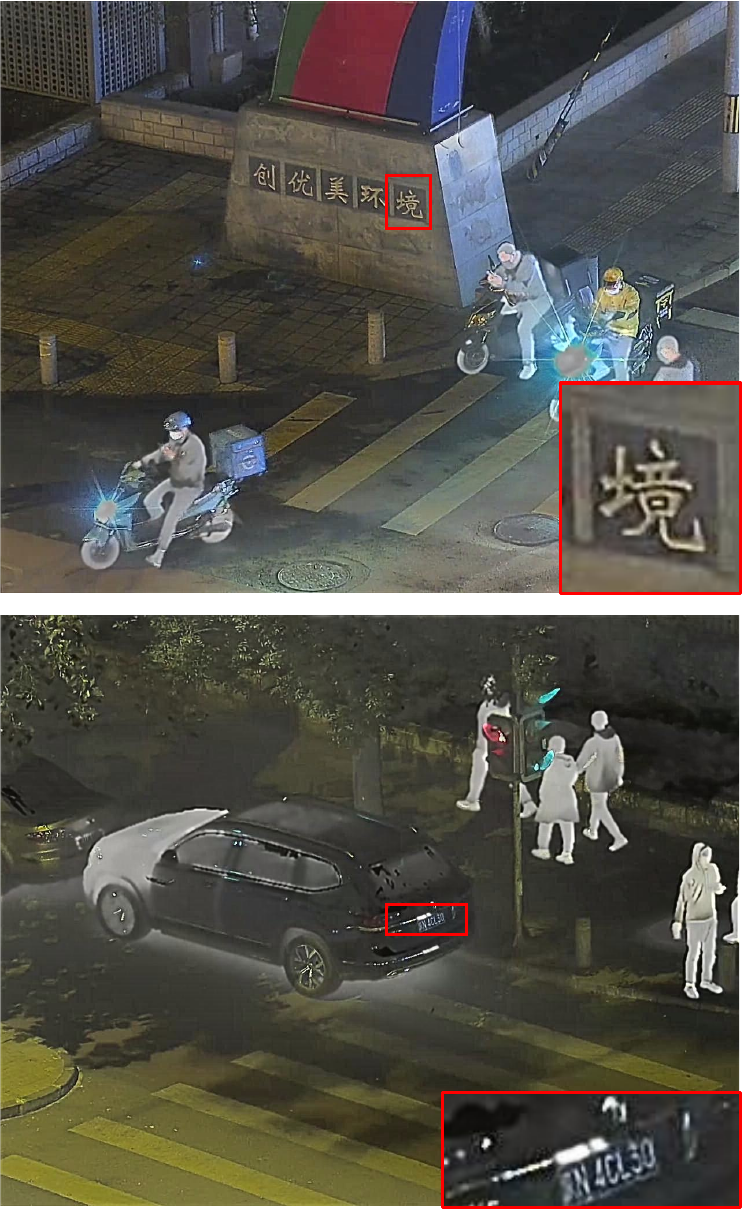}}
\vspace{5pt}
\caption{Vision quality comparison of five SOTA fusion methods on the LLVIP dataset. (a)-(b) Source images. (c) SeAFusion\cite{tang2022image}. (d) U2Fusion\cite{xu2020u2fusion}. (e) DenseFuse\cite{li2018densefuse}. (f) TarDAL\cite{liu2022target}. (g) DFVO. }
\vspace{-15pt}
\label{fig:results_1}
\end{figure*}

For the infrared image-reconstruction task, the purpose of the network is to generate infrared images suitable for the image fusion task. The loss function for this task can be defined as:
\begin{equation}
\mathcal{L}_{1} = \Vert{I^{ir} - \widetilde{I}^{ir}}\Vert_{1},
\end{equation}
where $\widetilde{I}^{ir}$ denotes the original infrared image. $I^{ir}$ represents the reconstructed infrared image. $\Vert \cdot \Vert_{1}$ represents the L1-norm.

Whereas the purpose of the illumination disentanglement task is to strip the degenerative illumination, a decomposition loss function consisting of four terms is designed:
\begin{equation}
\begin{aligned}
\mathcal{L}_{2} &= \lambda_1\mathcal{L}_{retinex} \\ &+ \lambda_2\mathcal{L}_{mc} + \lambda_3\mathcal{L}_{smooth} + \lambda_4\mathcal{L}_{his},
\label{loss: decom}
\end{aligned}
\end{equation}
where $\lambda_1$, $\lambda_2$, $\lambda_3$, and $\lambda_4$ represent the hyper-parameters that control the weights of corresponding item. $\mathcal{L}_{retinex}$ represents the restoration loss, which can guide the separation of degraded illumination according to the Retinex theory. The formula is as follows:
\begin{equation}
\mathcal{L}_{retinex} = \Vert I^{R} \cdot I^{il} - I^{y} \Vert_{1}.
\end{equation}
According to the analysis of illumination in KinD \cite{zhang2019kindling} and RetinexNet \cite{wei2018deep}, $\mathcal{L}_{mc}$ and $\mathcal{L}_{smooth}$ can further guide the network to generate the illumination image $I^{il}$ from the features, represented as:
\begin{equation}
\begin{aligned}
\mathcal{L}_{smooth} &= \left \Vert \frac{|\nabla I^{il}|}{\max(|\nabla I^{y}|, \epsilon)} \right \Vert_{1}, \\
\mathcal{L}_{mc} &= \Vert \, |\nabla I^{il}| \cdot \exp(-c \cdot |\nabla I^{il}|) \, \Vert_{1}, 
\end{aligned}
\end{equation}
where $\nabla$ represents the gradient operation. $|\cdot|$ represents the absolute value operation. $c$ denotes a parameter (set to 10) that controls the gradient values of the image. $\epsilon$ acts as the constant term (set to 0.01) to prevent division by $0$ in the denominator.

Since the ground-truth dataset is not available in unsupervised learning, resulting in a lack of guided structural information in the enhanced image, we adopt a loss function based on histogram equalization prior\cite{zhang2023self}, which is defined as follows:
\begin{equation}
\mathcal{L}_{his} = \left \Vert \mathcal{F}(I^{his}_{RGB}) - \mathcal{F}(I^{R}_{RGB}) \right \Vert_{2},
\end{equation}
where $I^{his}_{RGB}$ denotes the visible image $\widetilde{I}^{vi}$ after histogram equalization, which in the RGB space. $I^{R}_{RGB}$ represents the enhanced image on the RGB channels. $\Vert \cdot \Vert_{2}$ represents the L2-norm. $\mathcal{F}$ represents the feature extraction network, which consists of a pre-trained model of ResNet-18\cite{he2016deep}.

To achieve illumination disentanglement in the fused image, we utilize the enhanced and infrared images generated by the image-reconstruction and disentanglement tasks to guide feature information fusion as follows:
\begin{equation}
\mathcal{L}_{3}=\alpha_1\mathcal{L}_{cont}+\alpha_2\mathcal{L}_{str}+\alpha_3\mathcal{L}_{cos}, 
\label{loss: fusion}
\end{equation}
where $\mathcal{L}_{cont}$ represents the content loss, which is used to learn the texture information from multi-modal images. $\mathcal{L}_{str}$ is a structural loss, it can constrain the fused image to preserve the significant structural information of multi-modal images. $\mathcal{L}_{cos}$ denotes the color consistency loss, which is used to guide the correct color representation of the fused image (RGB space). $\alpha_1$, $\alpha_2$, and $\alpha_3$ are hyper-parameters that maintains the balance in each item. 

\begin{algorithm}[!t]
\caption{Training procedure of DFVO.}
\label{alg:alg1}
\renewcommand{\algorithmicrequire}{\textbf{Input:}}
\renewcommand{\algorithmicensure}{\textbf{Output:}}
\begin{algorithmic}[1]
\REQUIRE Visible images $\widetilde{I}^{vi}$ and infrared images $\widetilde{I}^{ir}$
\ENSURE Fused images $I^f_{RGB}$
\FOR{$n \le Max$ iterations $N$} 
\STATE Select $b$ visible images $\{ \widetilde{I}^{vi}_{1}, \widetilde{I}^{vi}_{2}, \cdots, \widetilde{I}^{vi}_{b}\};$
\STATE Extract the Y-channel images $\{I^y_1, I^y_2, \cdots, I^y_b\};$
\STATE Select $b$ infrared images $\{ \widetilde{I}^{ir}_{1}, \widetilde{I}^{ir}_{2}, \cdots, \widetilde{I}^{ir}_{b}\};$
\STATE Update the weight of total loss $\zeta_t$ in Eq. (\ref{equ: total loss});
\STATE Update the parameters of the network $\mathcal{N}$ by
Adam Optimizer: $\nabla_{\mathcal{N}}(\mathcal{L}_{\theta}(\mathcal{N}))$;
\STATE Obtain the fused images: $I_{RGB}^f \leftarrow \{ I^f, I^{cr}, I^{cb} \}$;
\ENDFOR
\end{algorithmic}
\end{algorithm}

In order to mitigate the impact of local over-exposed pixels on the fused image, we preprocess the enhanced image before inputting it into the fusion task loss. The specific expression is as follows:
\begin{equation}
\hat{I}^{R} = \sum\limits_{i=1}^N{\sigma_{R}I^R_i + \sigma_{ir}I^{ir}_i},
\label{equ: new enhanced}
\end{equation}
where $N$ is the total number of pixels. $i$ is the $i$-th pixel. $\sigma_{R}$ and $\sigma_{ir}$ represent the filter weights, whose formulated form is as follows:
\begin{equation}
\begin{cases}
\sigma_{R}=1 \, \& \, \sigma_{ir}=0, & I^{min}_i \le \eta \cdot mean(I^{min})\\
\sigma_{R}=0 \, \& \, \sigma_{ir}=1, & I^{min}_i > \eta \cdot mean(I^{min}),
\end{cases}
\end{equation}
among them, $\eta$ represents the parameter (set to 7) used to adjust the scale, allowing over-exposed pixels to be properly stripped. $I^{min}$ denotes finding the minimum value of RGB pixels in the image, expressed as:
\begin{equation}
I^{min}={\min}_{y\in \mathcal{W}(i)}({\min}_{c\in(R,G,B)}I_{y,c}^R),
\end{equation}
where $\mathcal{W}(i)$ represents the adjacent pixels of $i$ in its $10 \times 10$ window. $c$ denotes the channel. In the above process, through extensive experiments\cite{he2010single}, it has been observed that $I^{min}_i$ for certain pixels, such as those found in traffic lights, tends to approach $0$, while the values of locally over-exposed pixels on the RGB channels are relatively large. By comparing with the mean value of $I^{min}$, it is possible to approximately identify locally over-exposed pixels.

The content loss $\mathcal{L}_{cont}$ forces the network to learn more texture details from corrected-enhanced and infrared images, which can be defined as:
\begin{equation}
\mathcal{L}_{cont} = \omega_{R} \cdot \Vert I^{f} - \hat{I}^{R} \Vert_{2} + \omega_{ir} \cdot \Vert I^{f} - I^{ir} \Vert_{2},
\end{equation}
where $I^f$ is the fused image. $\omega_{R}$ and $\omega_{ir}$ are utilized to adaptively guide the image fusion, maintaining the optimal fusion between the multi-modal images. The specific process can be formulated as follows:
\begin{equation}
[ \, \omega_{R}, \omega_{ir} \, ] = softmax([ \, \mathbb{V}(\hat{I}^{R}), \mathbb{V}(I^{ir}) \, ]),
\end{equation}
where $\mathbb{V}$ represents the feature extraction operation, which is defined as:
\begin{equation}
\mathbb{V}(I)= \frac{1}{HW} \Vert \nabla(I) \Vert_{1},
\end{equation}
where $H$ and $W$ denote the height and
width of the input tensor $I$. $\nabla(\cdot)$ represents the Sobel operation. As for the structural loss $\mathcal{L}_{str}$, its core function is to guide the network to learn the structural information. The specific expression is:
\begin{equation}
\mathcal{L}_{str} = \Vert \, |\nabla I^f| - \max(|\nabla \hat{I}^{R}|, |\nabla I^{ir}|) \, \Vert_{2},
\end{equation}
where $|\cdot|$ represents the absolute value
operation. Considering that the fused image should visually approximate the visible image\cite{tang2023divfusion}, we utilize consistency loss to constrain the fused image on the RGB channels, defined as:
\begin{equation}
\mathcal{L}_{cos} = \frac{1}{HWC} \sum\limits_{c \in (R,G, B)}^{C} {cos(I^{f}_{RGB}, I^{R}_{RGB})},
\end{equation}
where $C$ indicates total numbers of the R,G,B channels. $cos(\cdot, \cdot)$ illustrates the discrete cosine similarity.

It is worth noting that the image fusion task as the primary task in the cascaded multi-task learning, its task-determining parameter should be gradually adjusted. To this end, we define the $\zeta_3$ similar to learning rate guidance. The specific expression is:
\begin{equation}
\zeta_3 = \mu \cdot (n - 1).
\label{equ:guidance_parameter}
\end{equation}
Here, $\mu$ indicates the parameter for preventing excessively large values. $n$ denotes the $n$-th iteration.

\section{Experiments}
\label{sec:ep}


\begin{figure}[!t]
\centering
\includegraphics[width=0.95\linewidth]{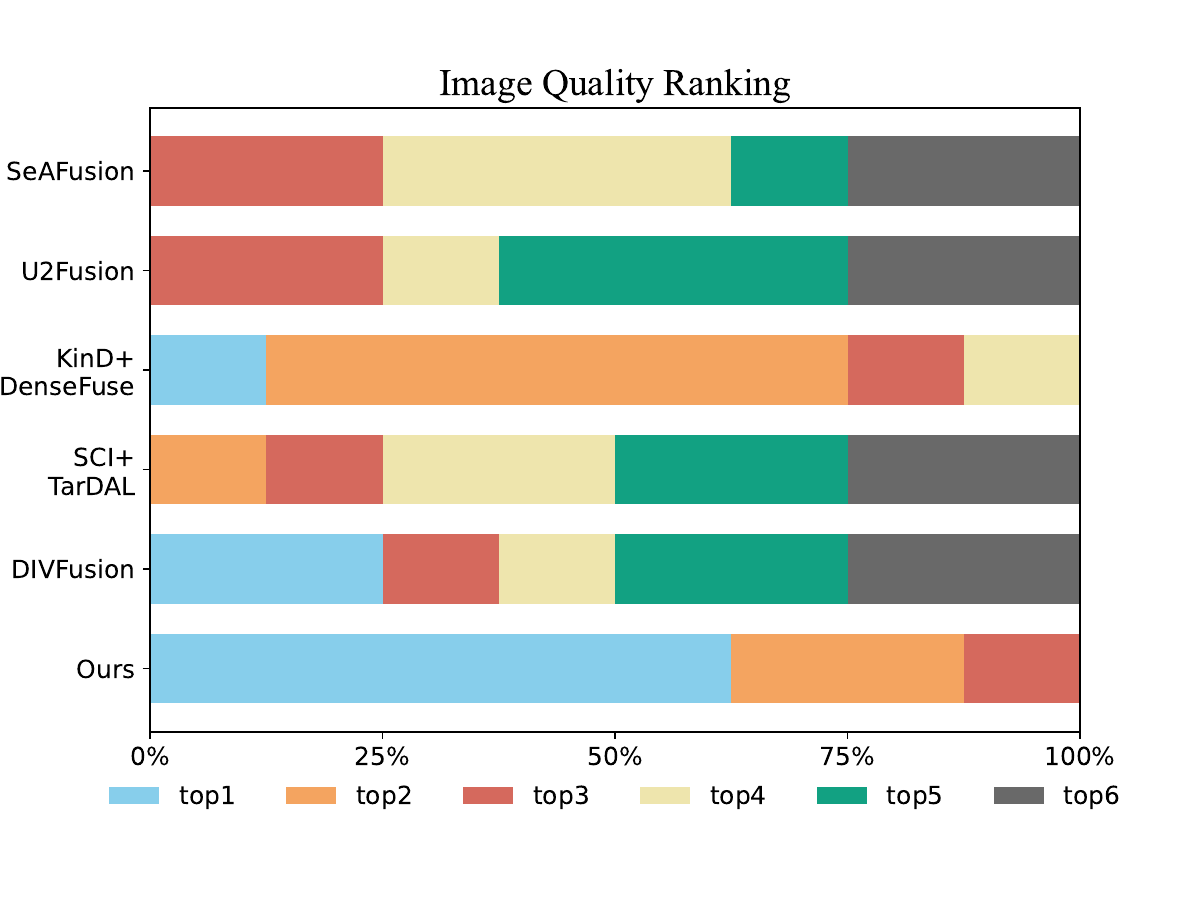}
\vspace{-4mm}
\caption{Human perception comparison of five SOTA fusion methods on the LLVIP dataset.}
\label{fig:mos}
\end{figure}

\subsection{Experimental Configurations}

\emph{1) Dataset:} We select the LLVIP dataset~\cite{jia2021llvip} to train our DFVO and other comparative fusion methods. During the training phase, we utilize 240 pairs of nighttime visible and infrared images to train our method. Among them, we randomly crop the image pairs to a patch size of $256 \times 320$ and shuffle the order of image pairs in the dataset before inputting them into the network, aiming to increase the training sample data.  It is worth noting that all images are constrained between 0 and 1 through normalization. Regarding the validation process, we use 50 typical visible and infrared images pairs from the LLVIP dataset to comprehensively demonstrate the superiority of the proposed method. For real-world testing, we conduct qualitative and quantitative experiments on 183, 331, and 25 pairs of images selected from the MSRS, SMOD, and KAIST datasets, respectively. Additionally, all images from the testing dataset are used in the original spatial resolution to capture more detailed information.

\emph{2) Training Details:} The entire training process is shown in Algorithm \ref{alg:alg1}. In practice, we set the batchsize $b=1$, and the total number of training epoch $N=100$, respectively. Besides, we adopt the Adam optimizer with $\beta_{1}$ of 0.9, $\beta_{2}$ of 0.8, epsilon of $10^{-8}$, the initial learning rate of $10^{-4}$ to optimize our model with the guidance of loss function. In addition, $\zeta_t$ of Eq. (\ref{equ: total loss}) are set as follows: $\zeta_1 = 10$, $\zeta_2 = 1$, and the $\mu = 0.1$ of $\zeta_3$. $\lambda_{(\cdot)}$ of Eq. (\ref{loss: decom}) in this work are set as follows: $\lambda_1 = 500$, $\lambda_2 = 1.35$, $\lambda_3 = 1.55$, and $\lambda_4 = 2.5$. $\alpha_{(\cdot)}$ of Eq. (\ref{loss: fusion}) are set as follows: $\alpha_1 = 1.75$, $\alpha_2 = 0.65$, $\alpha_3 = 0.35$. All experimental work is trained with the Pytorch framework on NVIDIA RTX 3090 GPU and 3.3GHz Intel Core i9-10940X CPU.

\emph{3) Metrics:} In this work, we use nine metrics to evaluate our method and other methods, which are entropy (EN), peak signal to noise ratio (PSNR), mean square error (MSE), mutual information (MI), spatial frequency (SF), standard deviation (SD), visual information fidelity (VIF), average gradient (AG), and correlation coefficient (CC).
\begin{figure*}[!ht]
\centering
\subfloat[Visible Image]{
        \rotatebox{90}{\scriptsize{~~~~~~~~KinD~~~~~~~~~~~~~~~~~SCI~~~~~~~~~~~~~~~Zero-DCE}}
		\includegraphics[width=0.95in]{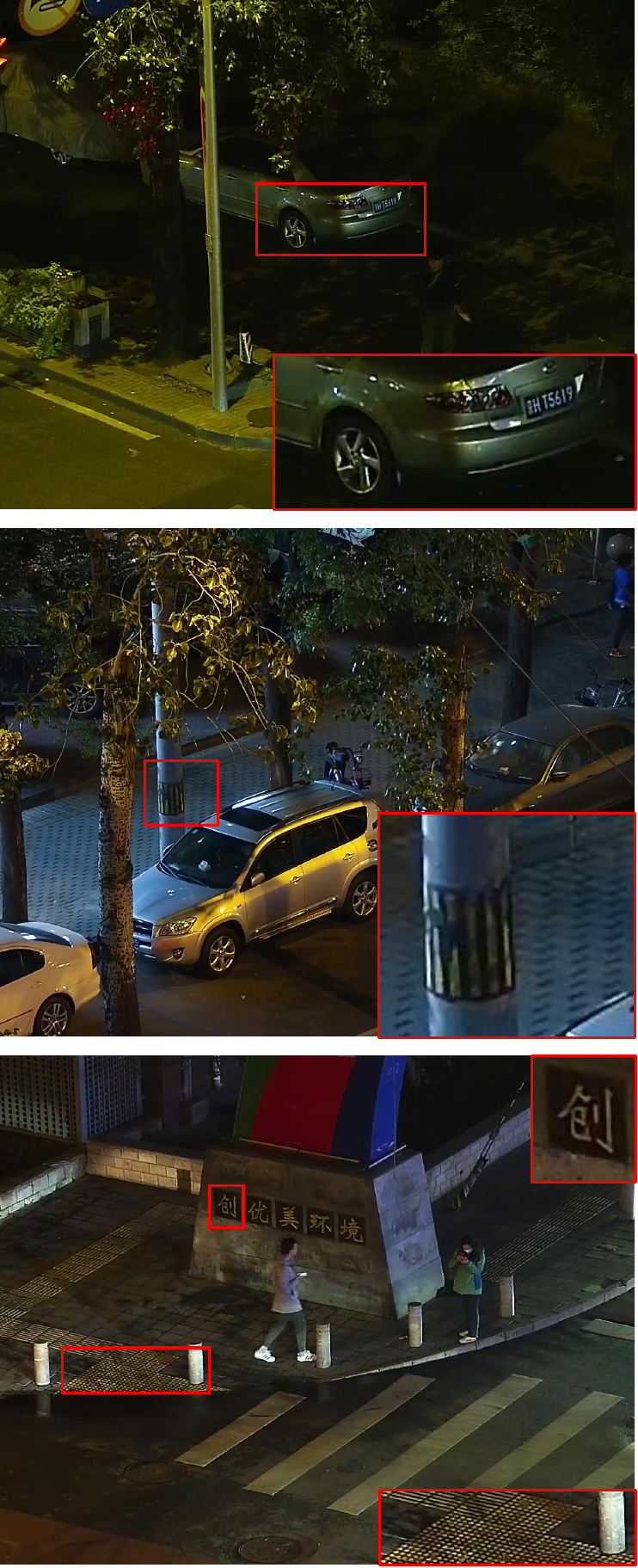}}
\subfloat[Infrared Image]{
		\includegraphics[width=0.95in]{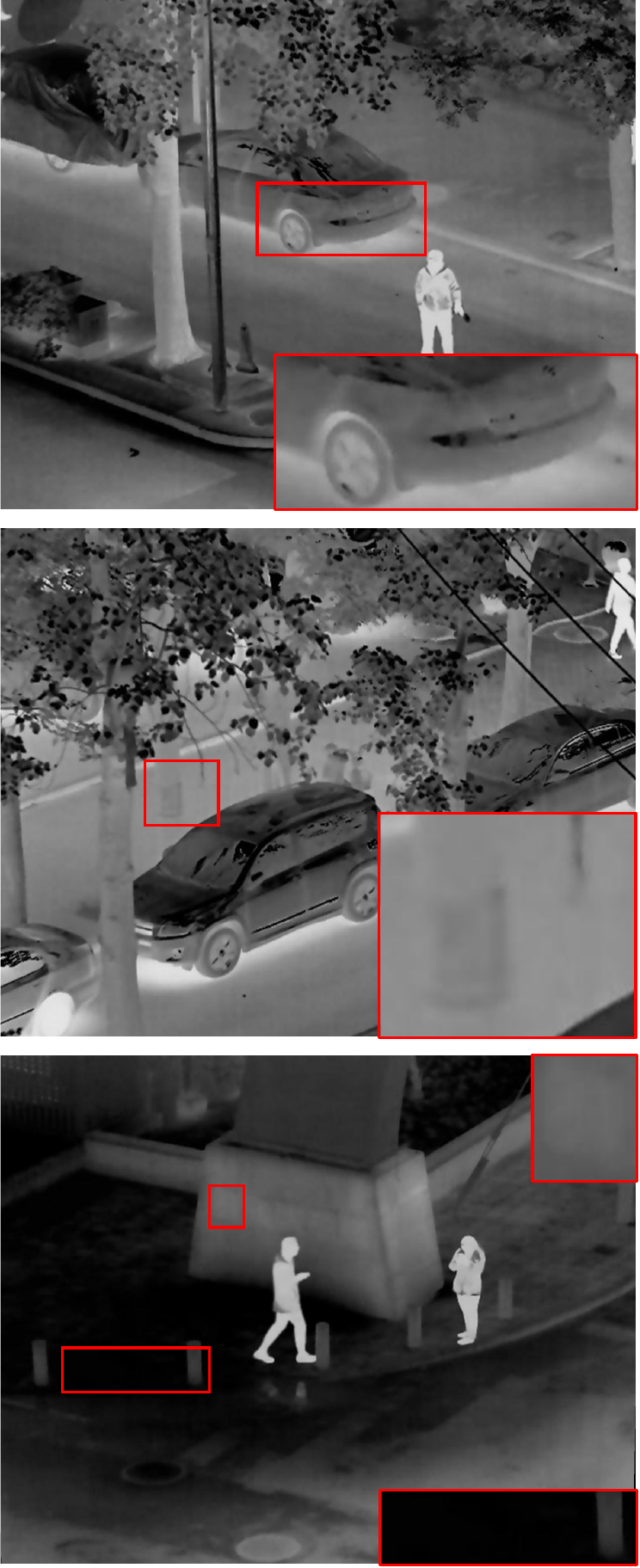}}
\subfloat[DenseFuse]{
		\includegraphics[width=0.95in]{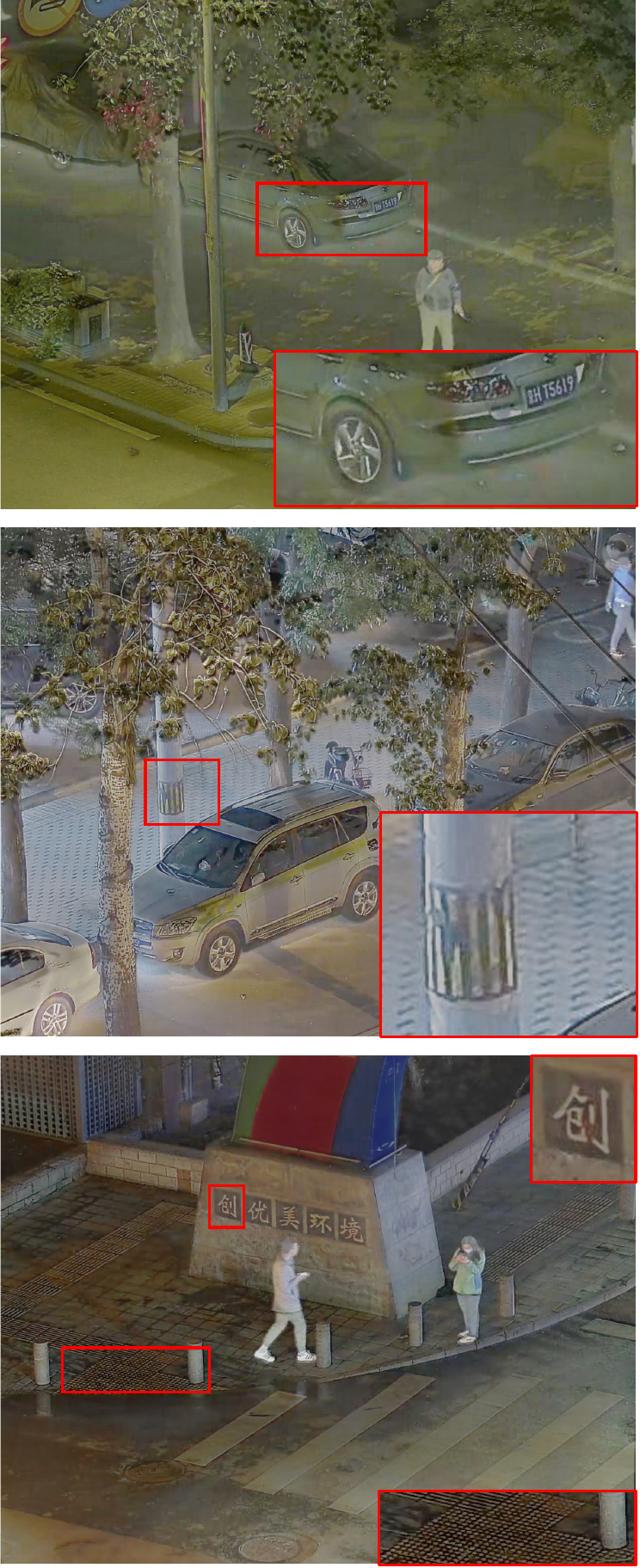}}
\subfloat[TarDAL]{
		\includegraphics[width=0.95in]{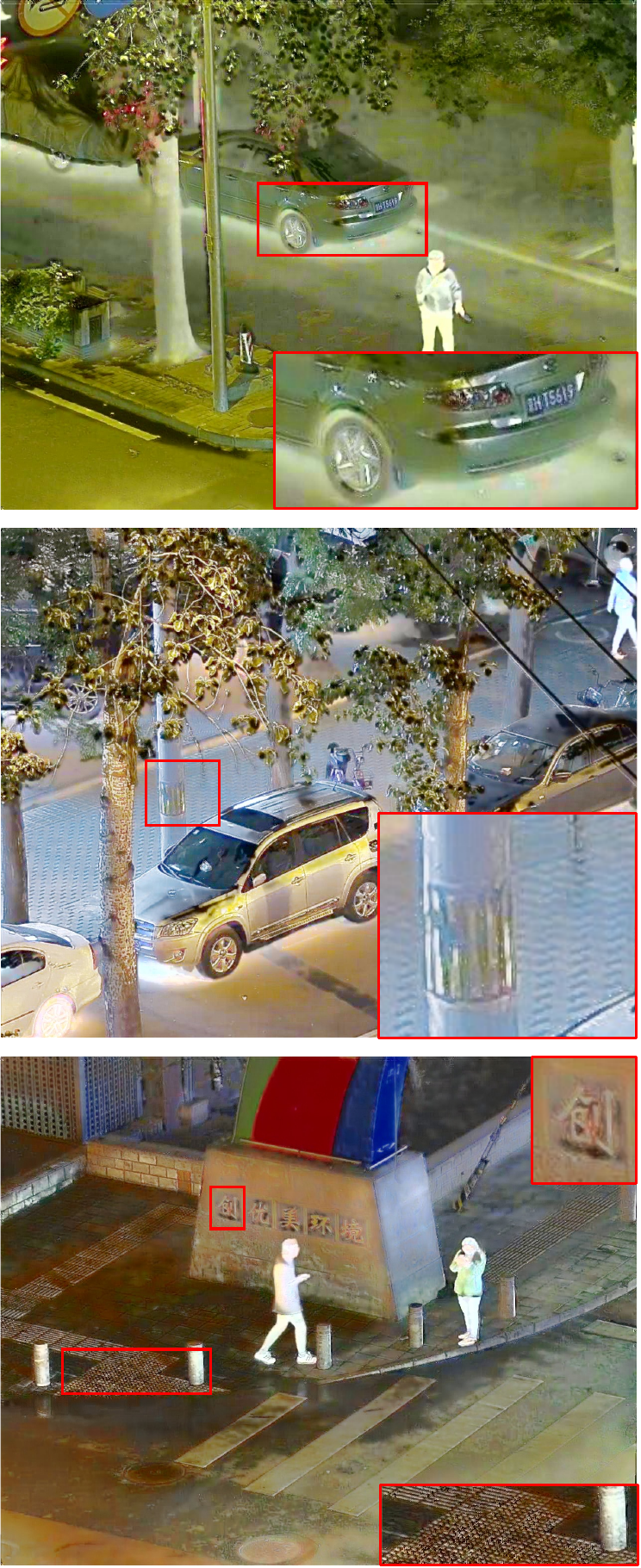}}
\subfloat[DIVFusion]{
		\includegraphics[width=0.95in]{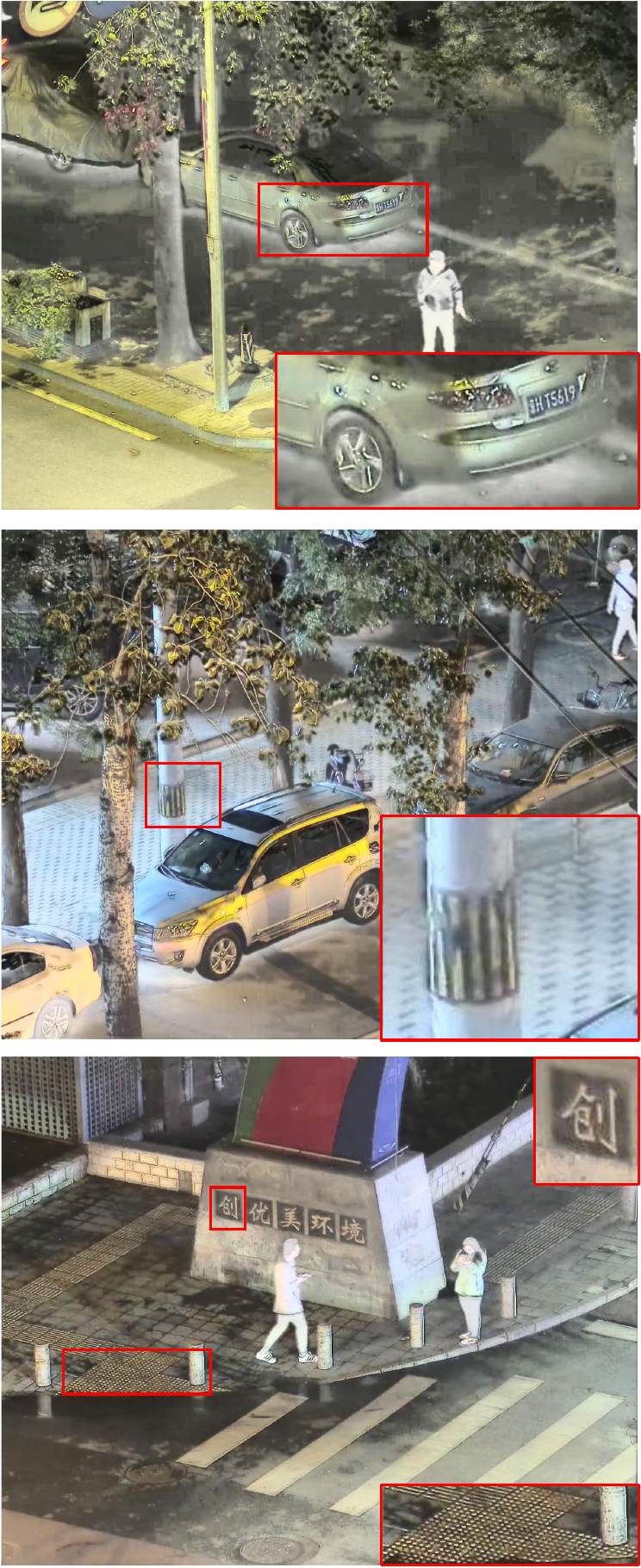}}
\subfloat[LENFusion]{
		\includegraphics[width=0.95in]{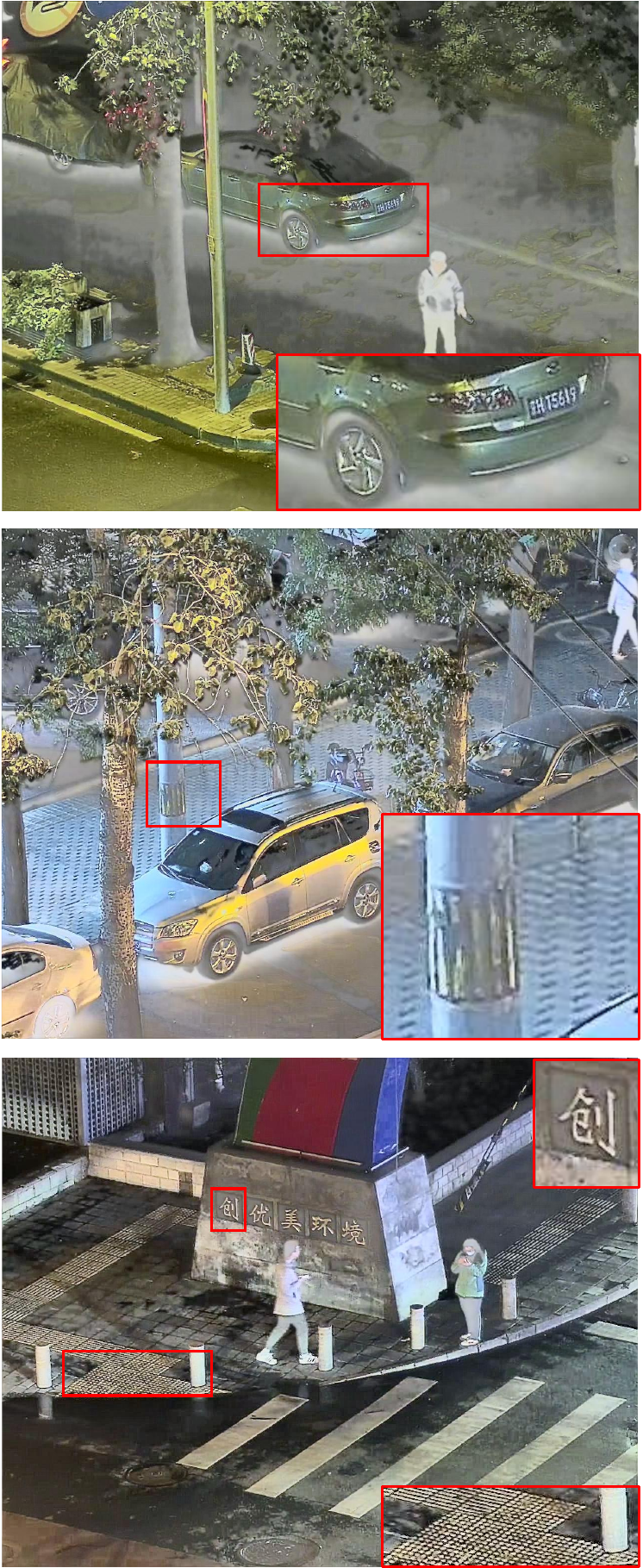}}
\subfloat[Ours]{
        \includegraphics[width=0.95in]{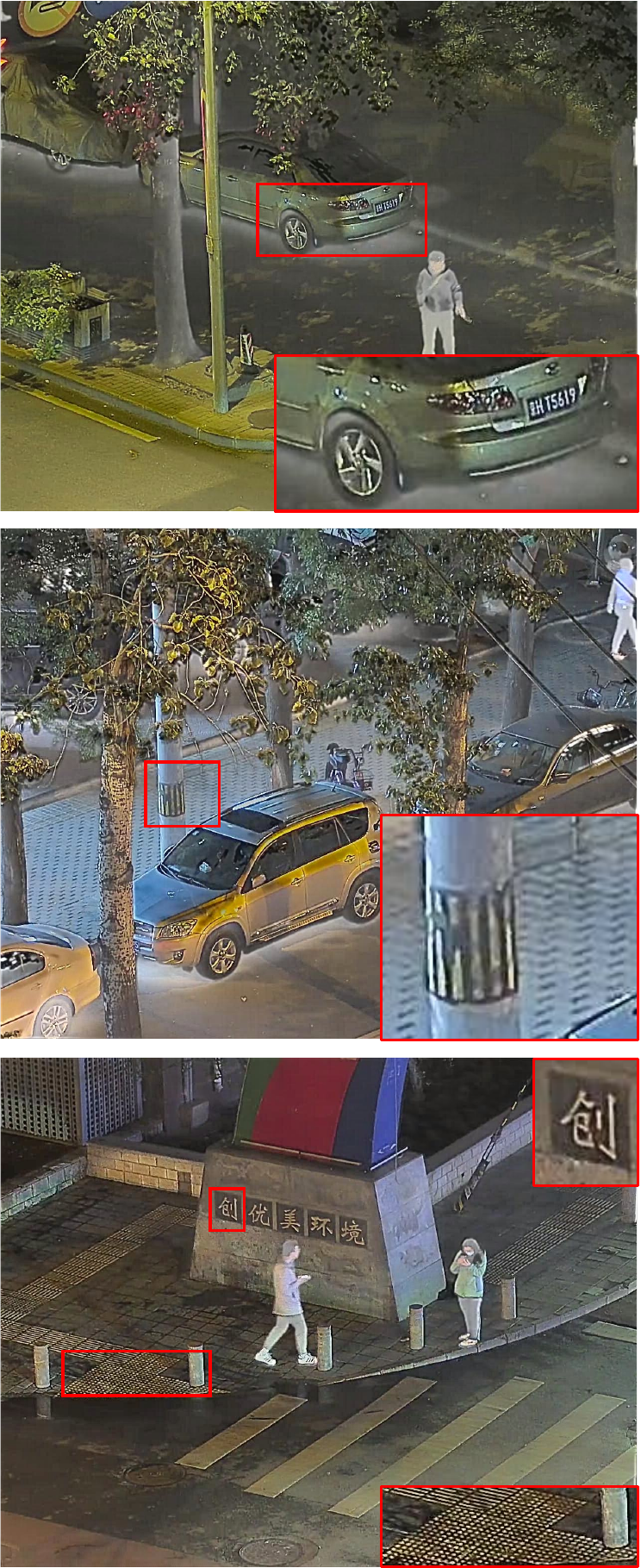}}
\vspace{5pt}
\caption{Vision quality comparison of two-stage fusion methods on the LLVIP dataset. (a)-(b) Source images. (c) Zero-DCE\cite{guo2020zero} / SCI\cite{ma2022toward} / KinD\cite{zhang2019kindling}+DenseFuse\cite{li2018densefuse}. (d) Zero-DCE\cite{guo2020zero} / SCI\cite{ma2022toward} / KinD\cite{zhang2019kindling}+TarDAL\cite{liu2022target}. (e) DIVFusion\cite{tang2023divfusion}. (f) LENFusion\cite{chen2024lenfusion}. (g) DFVO.\\
\textbf{( Two-stage fusion methods: Firstly, we employ the baseline low-light enhancement model to process the darkness visible image. Secondly, we apply the baseline image fusion method to fuse the enhanced visible image with the infrared image. )}}
\vspace{-15pt}
\label{fig:results_2}
\end{figure*}



\subsection{Performance Analysis on Image Fusion Tasks}
In this section, we compare our method with several SOTA image fusion algorithms to comprehensively demonstrate the advantages of our approach. Specifically, we select the proposed method with GANMcC\cite{ma2020ganmcc}, SeAFusion\cite{tang2022image}, U2Fusion\cite{xu2020u2fusion}, DenseFuse\cite{li2018densefuse}, TarDAL\cite{liu2022target}, DIVFusion\cite{tang2023divfusion}, and the latest published LENFusion\cite{chen2024lenfusion} to demonstrate the superiority of our method. In addition, we implement qualitative and quantitative comparisons with the two-stage fusion methods. 

\emph{1) Human Perception Analysis:} We analyze the image quality of each SOTA method from the perspective of the human visual perception. Specifically, We employ the mean opinion score (MOS) metric to quantify the perceptual quality. Firstly, we randomly select 50 images from the LLVIP dataset as the test data for the baseline model and generate fused images. Then, we invite 8 volunteers to participate in the testing process and ask them to subjectively select the best and worst fused images according to their preference. Finally, the scoring results are subjected to statistical analysis. As shown in Fig. \ref{fig:mos}, our DFVO outperforms others methods, obtaining a 62.5\% best perceptual experience. This indicates that our method can still produce good fusion results even in the presence of nighttime lighting contamination.

\emph{2) Qualitative Analysis:} To fully demonstrate the fusion performance of different algorithms, we select several diverse nighttime scenes, as shown in Fig. \ref{fig:results_1}. Due to the degradation caused by nighttime illumination, aggregating meaningful information from source images has become a challenge. As shown in the red box of images, the results of SeAFusion and TarDAL are missing a lot of texture detail information. U2Fusion and DenseFuse have more scene information, but the overall image appears visually dark. Although DIVFusion can enhance the overall brightness and preserve texture information hidden in the darkness, the fused image exhibits some undesirable phenomena, such as whiting, over-exposure, and edge blurring. By contrast, owing to the holistic network, the implementation of the cascaded multi-task strategy, and the meticulously designed loss function, our results showcase superior visual effects. 

In order to further illustrate the superiority of our approach, we conduct the comparison of visual quality with the two-stage fusion methods. As depicted in Fig. \ref{fig:results_2}, the two-stage fusion methods brighten the image but lose contrast and structure. DIVFusion has clarity and perception issues due to cascaded network information loss, and LENFusion, while good at brightness and gradient, has noise artifacts from its cascaded strategy. In summary, our model  disentangles illumination, learns source image features with less loss, and is better for high-level vision tasks.

\begin{table}[t]
\renewcommand{\arraystretch}{1.25}
\caption{
Quantitative results of the baseline fusion methods.
\label{tab:table1}}
\centering
\begin{tabularx}{\linewidth}{>{\centering\arraybackslash}c *{5}{>{\centering\arraybackslash}X}}
\hline
Method            & EN         & SF           & VIF               & AG              & CC        \\ \hline
GANMcC & 6.697      & 0.031        & 0.776             & 2.121           & \uline{0.718}     \\
DenseFuse & 6.657      & 0.029        & 0.766             & 2.163           & 0.712     \\
U2Fusion & 6.427      & 0.028        & 0.734             & 2.183           & 0.715     \\
TarDAL & \textbf{7.208}     & \uline{0.048}        & \uline{0.825}             & \uline{3.198}           & 0.651     \\
Ours             & \uline{7.172}      & \textbf{0.082}        & \textbf{0.961}             & \textbf{6.783}           & \textbf{0.724}     \\ \hline
\end{tabularx}
\vspace{-12pt}
\end{table}
\begin{table}[t]
\renewcommand{\arraystretch}{1.25}
\caption{
Quantitative results of two-stage fusion methods on the LLVIP dataset.}
\label{tab:table2}
\centering
\resizebox{\linewidth}{!}{
\begin{tabular}{cccccccccc}
\hline
\multicolumn{2}{c}{Method}                                & EN            & PSNR           & MSE            & SF            & SD             & VIF            & AG            & CC   \\ \hline
\multicolumn{1}{c|}{\multirow{3}{*}{Zero-DCE}} &SeAFusion & 7.032         & 58.445         & 0.092          & 0.065         & \textbf{9.242}          & 0.936          & 5.521         & 0.654  \\
\multicolumn{1}{c|}{}                          &DenseFuse & 6.773         & \uline{62.675}         & \uline{0.035}          & 0.036         & \uline{9.308}          & 0.779          & 3.061         & 0.709  \\
\multicolumn{1}{c|}{}                          &TarDAL    & 7.324         & 60.203         & 0.063          & 0.056         & 9.688          & 0.825          & 4.235         & 0.661  \\ \hline
\multicolumn{1}{c|}{\multirow{3}{*}{SCI}}      &U2Fusion  & 7.037         & 60.902         & 0.065          & 0.047         & 9.734          & 0.920          & 4.162         & 0.698  \\
\multicolumn{1}{c|}{}                          &DenseFuse & 6.982         & 62.283         & 0.039          & 0.049         & 9.611          & 0.909          & 4.035         & \uline{0.716}  \\
\multicolumn{1}{c|}{}                          &TarDAL    & \uline{7.499}         & 58.997         & 0.084          & 0.067         & 10.209         & 0.957          & 5.358         & 0.685  \\ \hline
\multicolumn{1}{c|}{\multirow{2}{*}{KinD}}     &DenseFuse & 6.846         & 62.658         & 0.039          & 0.036         & 9.556          & 0.881          & 3.127         & 0.699  \\
\multicolumn{1}{c|}{}                          &TarDAL   & 7.375         & 59.879         & 0.069          & 0.057         & 10.179         & 0.901          & 4.471         & 0.667  \\ \hline
\multicolumn{2}{c}{DIVFusion}                             & \textbf{7.587}         & 57.891         & 0.111          & 0.054         & 10.498        & \textbf{1.198}          & 4.501         & 0.675  \\
\multicolumn{2}{c}{LENFusion}                             & 7.437         & 59.943                  & 0.069          & \uline{0.079}        & 9.902                  & 0.958                    & \uline{6.442}         & 0.653  \\ \hline
\multicolumn{2}{c}{Ours}                                  & 7.172         & \textbf{63.258}         & \textbf{0.032}          & \textbf{0.082}         & 9.413          & \uline{0.961}          & \textbf{6.783}         & \textbf{0.724}  \\ \hline
\end{tabular}}
\vspace{-12pt}
\end{table}
\emph{3) Quantitative Analysis:} Table \ref{tab:table1} represents the quantitative results of the SOTA fusion methods. As shown in the results, our method achieves the best performance in terms of SF, VIF, AG, and CC, indicating that the fusion performance of our DFVO in low-light environments is superior. Table \ref{tab:table2} indicates the quantitative comparison results with two-stage fusion methods. Our DFVO achieves the best results in other five metrics and the second place in the VIF. The superior values in PSNR, MSE, SF, AG, and CC indicate that our fusion images possess less noise, clearer texture details, and richer gradient information. Specifically, DIVFusion tends to produce brighter luminance, which can easily incorporate information into the original visible image, including potential noise. 

\subsection{Generalization Analysis on Image Fusion Tasks}

To assess the generalization ability of DFVO, additional comparative experiments are conducted on the MSRS, SMOD, and KAIST datasets.

\begin{figure*}[ht]
\centering
\includegraphics[width=0.9\linewidth]{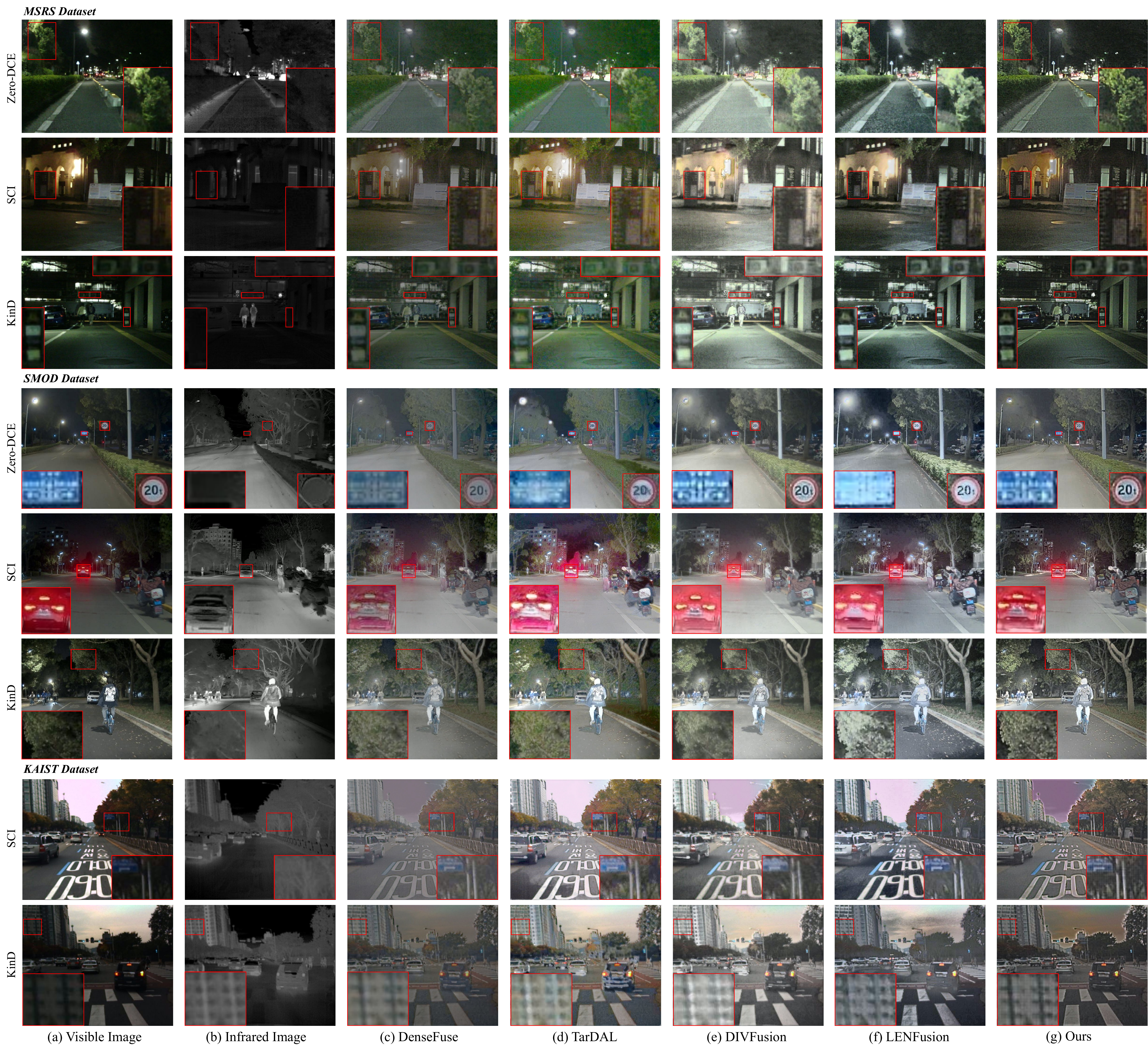}
\caption{Vision quality comparison of two-stage fusion methods on the MSRS, SMOD, and KAIST datasets. (a)-(b) Source images. (c) Zero-DCE\cite{guo2020zero} / SCI\cite{ma2022toward} / KinD\cite{zhang2019kindling}+DenseFuse\cite{li2018densefuse}. (d) Zero-DCE\cite{guo2020zero} / SCI\cite{ma2022toward} / KinD\cite{zhang2019kindling}+TarDAL\cite{liu2022target}. (e) DIVFusion\cite{tang2023divfusion}. (f) LENFusion\cite{chen2024lenfusion}. (g) DFVO.}
\label{fig:generalization_results}
\end{figure*}

\begin{table*}[!t]
\renewcommand{\arraystretch}{1.2}
\caption{
Quantitative results of the two-stage fusion methods across three different datasets.
\label{tab:generalization_table}}
\centering
\small
\begin{tabularx}{\linewidth}{*{10}{>{\centering\arraybackslash}X}}
\hline
\multicolumn{1}{c}{\multirow{3}{*}{Metric}} &\multicolumn{9}{c}{\textbf{MSRS Dataset}} \\ \cline{2-10}
\multicolumn{1}{c}{} &\multicolumn{2}{c|}{KinD} &\multicolumn{2}{c|}{SCI} &\multicolumn{2}{c}{Zero-DCE} &\multicolumn{1}{c}{\multirow{2}{*}{DIVFusion}} &\multicolumn{1}{c}{\multirow{2}{*}{LENFusion}} &\multicolumn{1}{c}{\multirow{2}{*}{Ours}} \\ \cline{2-7}
\multicolumn{1}{c}{} &\multicolumn{1}{c}{DenseFuse} &\multicolumn{1}{c|}{TarDAL} &\multicolumn{1}{c}{DenseFuse} &\multicolumn{1}{c|}{TarDAL} &\multicolumn{1}{c}{DenseFuse} &\multicolumn{1}{c}{TarDAL} &\multicolumn{1}{c}{} &\multicolumn{1}{c}{} &\multicolumn{1}{c}{} \\ \hline
SF & 0.020 & 0.037 & 0.028 & 0.044 & 0.023 & 0.041 & \uline{0.045} & 0.043 & \textbf{0.046} \\
PSNR &64.754 &59.690 &64.591 &58.786 &\uline{64.782} &59.846 &55.383 &61.376 &\textbf{65.038} \\
MSE  &0.024	 &0.071	 &\uline{0.023}	 &0.087	 &0.024	 &0.068	 &0.189	 &0.051	 &\textbf{0.022} \\
MI   &2.154	 &2.157	 &2.169	 &2.174	 &2.158	 &2.196	 &\textbf{2.235} &2.154	 &\uline{2.201} \\
CC   &0.587	 &0.579	 &0.580	 &0.561	 &\uline{0.588}	 &0.522	 &0.498	 &0.560	 &\textbf{0.594} \\
AG   &2.044	 &2.256	 &2.556	 &3.923	 &2.282	 &3.581	 &\uline{4.502}	 &4.182	 &\textbf{4.770} \\ \hline
\multicolumn{1}{c}{\multirow{3}{*}{Metric}} &\multicolumn{9}{c}{\textbf{SMOD Dataset}} \\ \cline{2-10}
\multicolumn{1}{c}{} &\multicolumn{2}{c|}{KinD} &\multicolumn{2}{c|}{SCI} &\multicolumn{2}{c}{Zero-DCE} &\multicolumn{1}{c}{\multirow{2}{*}{DIVFusion}} &\multicolumn{1}{c}{\multirow{2}{*}{LENFusion}} &\multicolumn{1}{c}{\multirow{2}{*}{Ours}} \\ \cline{2-7}
\multicolumn{1}{c}{} &\multicolumn{1}{c}{DenseFuse} &\multicolumn{1}{c|}{TarDAL} &\multicolumn{1}{c}{DenseFuse} &\multicolumn{1}{c|}{TarDAL} &\multicolumn{1}{c}{DenseFuse} &\multicolumn{1}{c}{TarDAL} &\multicolumn{1}{c}{} &\multicolumn{1}{c}{} &\multicolumn{1}{c}{} \\ \hline
SF	&0.030	&0.055	&0.038	&\uline{0.056}	&0.029	&0.052	&0.047	&0.052	&\textbf{0.061} \\
PSNR &62.003 &61.717 &61.815	&60.265	&\uline{62.242}	&61.475	&59.043	&60.433	&\textbf{62.327} \\
MSE	&0.041	&0.045	&0.043	&0.063	&\uline{0.039}	&0.047	&0.082	&0.059	&\textbf{0.038} \\
MI	&3.111	&3.246	&3.190	&3.134	&3.231	&\uline{3.257}	&3.235	&2.712	&\textbf{3.372} \\
CC	&0.806	&0.774	&\uline{0.823}	&0.795	&0.811	&0.773	&0.816	&0.748	&\textbf{0.849} \\
AG	&2.616	&4.089	&3.059	&4.725	&2.441	&3.880	&3.468	&\textbf{5.339}	&\uline{5.255} \\ \hline
\multicolumn{1}{c}{\multirow{3}{*}{Metric}} &\multicolumn{9}{c}{\textbf{KAIST Dataset}} \\ \cline{2-10}
\multicolumn{1}{c}{} &\multicolumn{3}{c|}{KinD} &\multicolumn{3}{c}{SCI} &\multicolumn{1}{c}{\multirow{2}{*}{DIVFusion}} &\multicolumn{1}{c}{\multirow{2}{*}{LENFusion}} &\multicolumn{1}{c}{\multirow{2}{*}{Ours}} \\ \cline{2-7}
\multicolumn{1}{c}{} &\multicolumn{1}{c}{DenseFuse} &\multicolumn{2}{c|}{TarDAL} &\multicolumn{1}{c}{DenseFuse} &\multicolumn{2}{c}{TarDAL} &\multicolumn{1}{c}{} & \multicolumn{1}{c}{} \\ \hline
SF	&0.026  &\multicolumn{2}{c}{0.042}	&0.034	&\multicolumn{2}{c}{\uline{0.053}}	&0.042	&0.047	&\textbf{0.060} \\
PSNR	&\uline{61.251}	&\multicolumn{2}{c}{58.032}	&60.168	&\multicolumn{2}{c}{57.284}	&56.842	&59.249	&\textbf{61.585} \\
MSE	&\uline{0.048}	&\multicolumn{2}{c}{0.102}	&0.063	&\multicolumn{2}{c}{0.122}	&0.135	&0.077	&\textbf{0.045} \\
MI	&1.804	&\multicolumn{2}{c}{\uline{2.333}}	&2.197	&\multicolumn{2}{c}{2.296}	&\textbf{2.492}	&2.096	&2.100 \\
AG	&2.736	&\multicolumn{2}{c}{5.032}	&3.118	&\multicolumn{2}{c}{5.556}	&4.667	&\uline{6.275}	&\textbf{6.361} \\ \hline
\end{tabularx}
\vspace{-15pt}
\end{table*}

\emph{1) Qualitative Analysis:} As illustrated in Fig. \ref{fig:generalization_results}, achieving a balance between fusion information and visibility continues to be a challenge. Two-stage fusion methods can enhance brightness but lead to loss of information, such as texture, contours, and some thermal targets. While DIVFusion achieves the highest brightness, it suffers from excessive noise and areas of overexposure in the fused images. By contrast, DFVO provides clearer texture details, including distant road signs and branches on the MSRS and SMOD datasets.

\emph{2) Quantitative Analysis:} As shown in Table \ref{tab:generalization_table}, DFVO leads in SF, PSNR, and MSR across various datasets, highlighting that the fused images generated by our method have reduced noise distribution, higher clarity, and richer detail. Additionally, our DFVO excels in CC and AG metrics on the MSRS dataset, indicating a stronger correlation with the source images and clearer texture details. On the SMOD dataset, our method performs well in MI and CC metrics, suggesting that DFVO can effectively integrate more information from the source images.

\emph{3) Computational Complexity Analysis:} 
To observe the computational complexity of models, we provide the number of parameters, floating-point operations(FLOPs) and running time for the aforementioned methods, as shown in Table \ref{tab:computational_table}. SCI and DenseFuse outperform in their domains. SCI leverages a highly lightweight network structure, whereas DenseFuse primarily emphasizes loss function and fusion strategies, maintaining minimal parameters during testing. Compared with darkness-free fusion, DFVO demonstrates a notable advantage. While the inclusion of the HCAM leads to more model parameters than others, DFVO excels in inference speed. Moreover, by removing the two-stage operation mechanism, it simplifies the model's deployment in autonomous driving applications.

\begin{table}[t]
\renewcommand{\arraystretch}{1.25}
\caption{
Quantitative results of the ablation study on 50 pairs of visible and infrared images from LLVIP dataset.
\label{tab:table4}}
\centering
\resizebox{\linewidth}{!} {
\begin{tabular}{lccccccc}
\hline
 &Configurations &EN &SF &AG &PSNR &MSE &CC \\ \hline
\uppercase\expandafter{\romannumeral1} &{CNN $\to$ DEM in LCFE} &7.045	&0.065	&5.113	&62.729	&0.037	&0.683 \\ 
\uppercase\expandafter{\romannumeral2} &{CNN $\to$ HCAM in LCFE} &7.144	&0.081	&6.641	&61.801	&0.043	&0.692 \\
\uppercase\expandafter{\romannumeral3} &{w/o $\omega_{R} \& \omega_{ir}$} &7.061 &\textbf{0.083}	&6.564	&62.125	&0.039	&0.662 \\
\uppercase\expandafter{\romannumeral4} &{w/o $\mathcal{L}_{cont}$} &6.510 &0.069 &4.592	&60.321	&0.060	&-0.029 \\
\uppercase\expandafter{\romannumeral5} &{w/o $\mathcal{L}_{str}$} &6.879 &0.038 &2.964	&63.231	&0.034	&0.716 \\
\uppercase\expandafter{\romannumeral6} &{w/o $\mathcal{L}_{cos}$} &7.115 &0.081 &6.284	&63.134	&0.036	&0.682 \\ \hline
 &Ours &\textbf{7.172}	&0.082	&\textbf{6.783}	&\textbf{63.268}	&\textbf{0.032}	&\textbf{0.724} \\ \hline
\end{tabular}}
\vspace{-12pt}
\end{table}

\subsection{Ablation Experiment}

\begin{figure}[!t]
\centering
\vspace{-6pt}
\subfloat[Visible Image]{
		\includegraphics[width=0.81in]{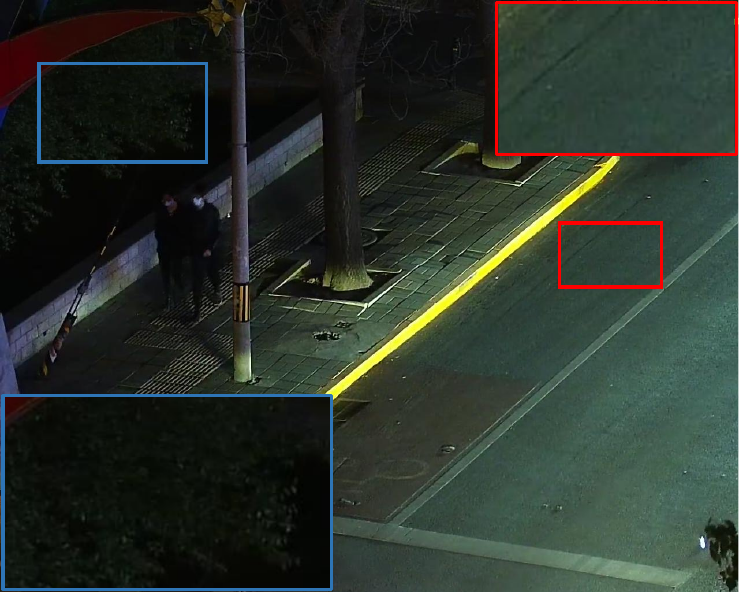}}
\subfloat[CNN$\to$DEM]{
		\includegraphics[width=0.81in]{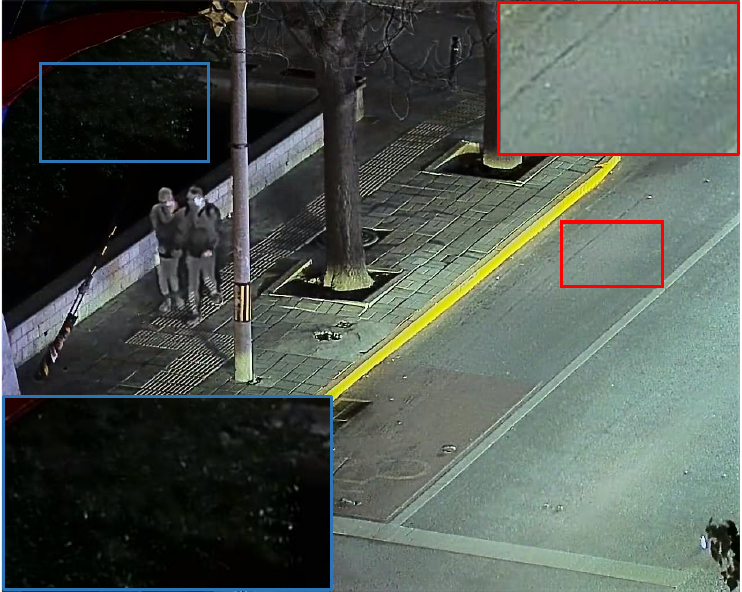}}
\subfloat[CNN$\to$HCAM]{
		\includegraphics[width=0.81in]{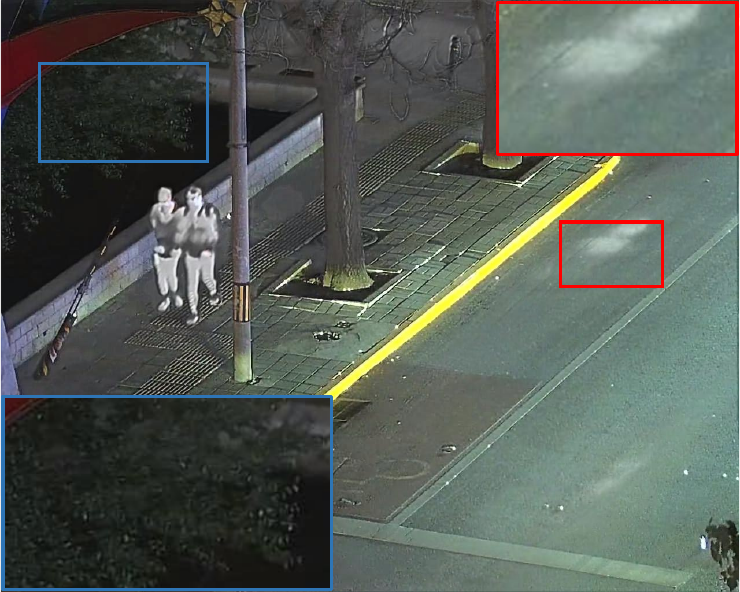}}
\subfloat[Ours]{
		\includegraphics[width=0.81in]{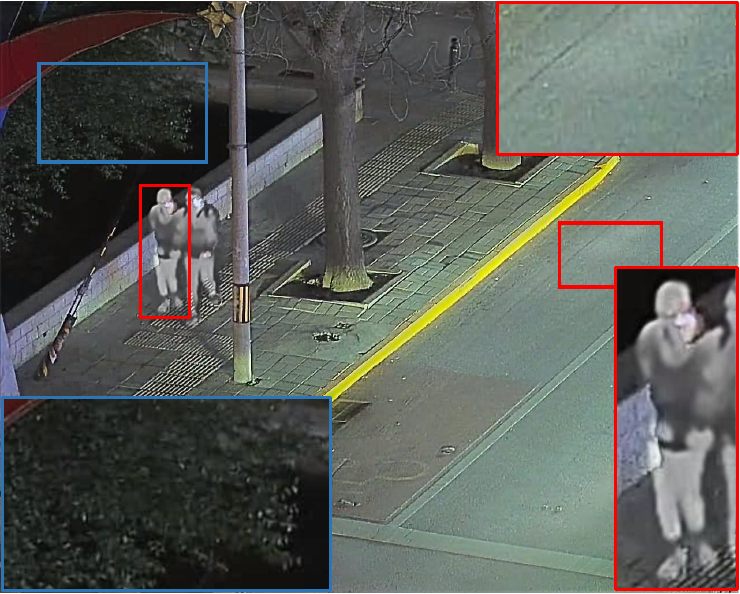}}
\\
\vspace{-2mm}
\subfloat[w/o $\omega_{R} \& \omega_{ir}$]{
		\includegraphics[width=0.81in]{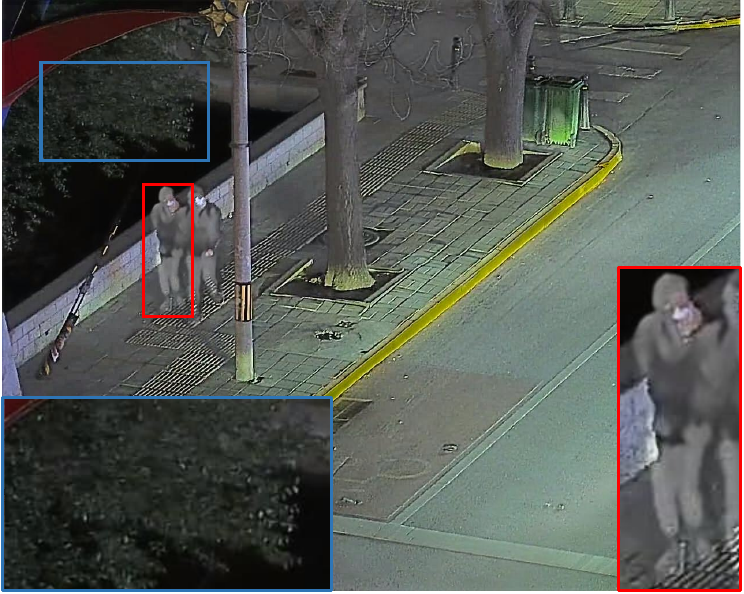}}
\subfloat[w/o $\mathcal{L}_{cont}$]{
		\includegraphics[width=0.81in]{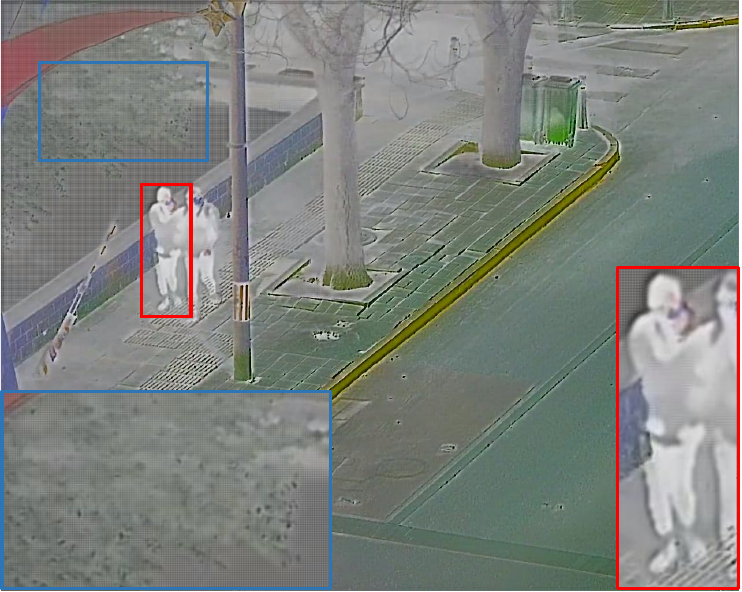}}
\subfloat[w/o $\mathcal{L}_{str}$]{
		\includegraphics[width=0.81in]{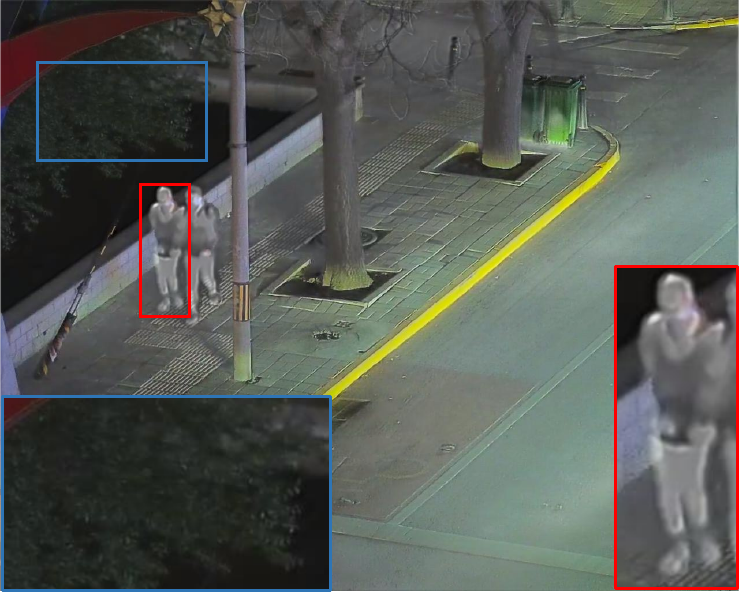}}
\subfloat[$\zeta_3$ = 10]{
		\includegraphics[width=0.81in]{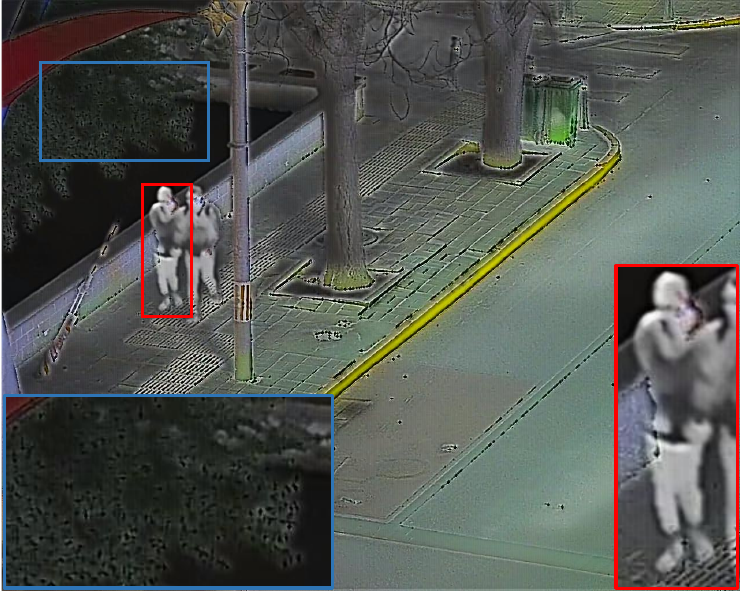}}
\vspace{5pt}
\caption{Vision quality comparison of the ablation study on important modules and loss functions.}
\vspace{-15pt}
\label{fig:ablation}
\end{figure}

\emph{1) Analysis of DEM:} The DEM is specifically designed to capture high-frequency information from the source images. In Exp. \uppercase\expandafter{\romannumeral1}, we change the INN blocks as CNN blocks. As illustrated in Fig. \ref{fig:ablation}(b) and Table \ref{tab:table4}, the ablation images exhibit a considerable loss of texture details, including leaves and branches. The quantitative experiments reveal a decline in EN, SF, and AG, which proves that the high-frequency information loss is serious when the fusion task is performed without DEM. 

\emph{2) Analysis of HCAM:} The HCAM is utilized to capture low-frequency features from the source images. Similarly, in Exp. \uppercase\expandafter{\romannumeral2}, we change the transformer modules as CNN blocks, the results are presented in Fig. \ref{fig:ablation}(c) and Table \ref{tab:table4}. In the absence of HCAM, our method experiences a loss of information in the fused image's global features, leading to a subpar visual experience. Simultaneously, metrics like PSNR, MSE, and CC also show a decline. Therefore, the utilization of HCAM in our LCFE plays a crucial role in generating visually pleasing fusion images.

\emph{3) Analysis of Loss Function:} The adaptive weights $\omega_{R} \& \omega_{ir}$ can adaptively guide the network to learn salient information from the source images. After removing the adaptive weights, the fusion images exhibit an abnormal contrast display issue. As shown in Fig. \ref{fig:ablation}(e), the headlights in the scene appear to vanish. Meanwhile, pedestrian targets on the street appear less salient compared to the complete DFVO. The content loss $\mathcal{L}_{cont}$ is one of the most important loss functions in our DFVO. As depicted in Fig. \ref{fig:ablation}(f), the fused results generated by the proposed method without the content loss exhibit poor visual quality with significant loss of texture, scene, and color information. The structural loss $\mathcal{L}_{str}$ guides the LCFE in the proposed network to acquire more structural information from the source images. As illustrated in Fig. \ref{fig:ablation}(g), the fused images lacking structural loss exhibit undesirable phenomena like edge blurring and loss of gradient information. The consistency loss $\mathcal{L}_{cos}$ ensures that the fused images maintain consistency with the visible images in terms of color information, thereby alleviating the color deviation phenomenon that may occur when directly converted to RGB space. The task-determining parameter $\zeta_3$ is specifically designed for implementing cascaded multi-task learning. As shown in Fig. \ref{fig:ablation}(h), we set $\zeta_3$ to 10 and retrain the model. The results reveal oscillations in the fused images, highlighting the significance of Eq. (\ref{equ:guidance_parameter}).

The quantitative analysis of the loss function is presented in Table \ref{tab:table4}. We can observe that our DFVO exhibits significant superiority in EN, AG, PSNR, MSE, and CC. The slight disadvantage in SF metrics suggests that although the detail extraction capability is marginally better without $\omega_{R} \& \omega_{ir}$, our DFVO proves to be superior overall.

\begin{table*}[!t]
\renewcommand{\arraystretch}{1.25}
\caption{
Computational comparison of the above comparison methods.
\label{tab:computational_table}}
\centering
\footnotesize
\begin{tabularx}{\linewidth}{*{10}{>{\centering\arraybackslash}X}}
\hline
\multicolumn{1}{c}{\multirow{2}{*}{Metric}} & \multicolumn{3}{c}{Low-Light Enhancement} & \multicolumn{3}{c}{Common Fusion} & \multicolumn{3}{c}{Darkness-Free Fusion} \\ \cline{2-10}
\multicolumn{1}{c}{} &\multicolumn{1}{c}{KinD} &\multicolumn{1}{c}{SCI} &\multicolumn{1}{c|}{Zero-DCE} &\multicolumn{1}{c}{GANMcC} &\multicolumn{1}{c}{DenseFuse} &\multicolumn{1}{c|}{TarDAL} &\multicolumn{1}{c}{DIVFusion} &\multicolumn{1}{c}{LENFusion} &\multicolumn{1}{c}{Ours} \\ \hline
Params(M) &- &3.48$\times 10^{-4}$ &0.080	&1.864	&0.074	&0.290	&4.403	&0.740	&9.573 \\ \hline
\multicolumn{10}{c}{1024 $\times$ 1280 Images as Inputs} \\ \cline{2-10}
FLOPs(G)  &- &0.700	&208.153 &4811.943 &230.961	&388.854 &14450.916	&2854.729 &3155.438 \\
RunTime(s) &5.471 &0.099 &0.230	&0.310	&0.515	&0.346	&1.505	&0.812	&0.654 \\ \hline
\multicolumn{10}{c}{480 $\times$ 640 Images as Inputs} \\ \cline{2-10}
FLOPs(G) &-	&0.164	&48.786	&1110.376	&54.132	&91.138	&3386.933	&669.047 &732.330 \\
RunTime(s) &1.363 &0.011 &0.140	&0.146	&0.407	&0.167	&0.415	&0.231	&0.186 \\ \hline
\end{tabularx}
\vspace{-15pt}
\end{table*}

\begin{figure}[t]
\centering
\vspace{-6pt}
\subfloat[Visible Image]{
		\includegraphics[width=1.1in]{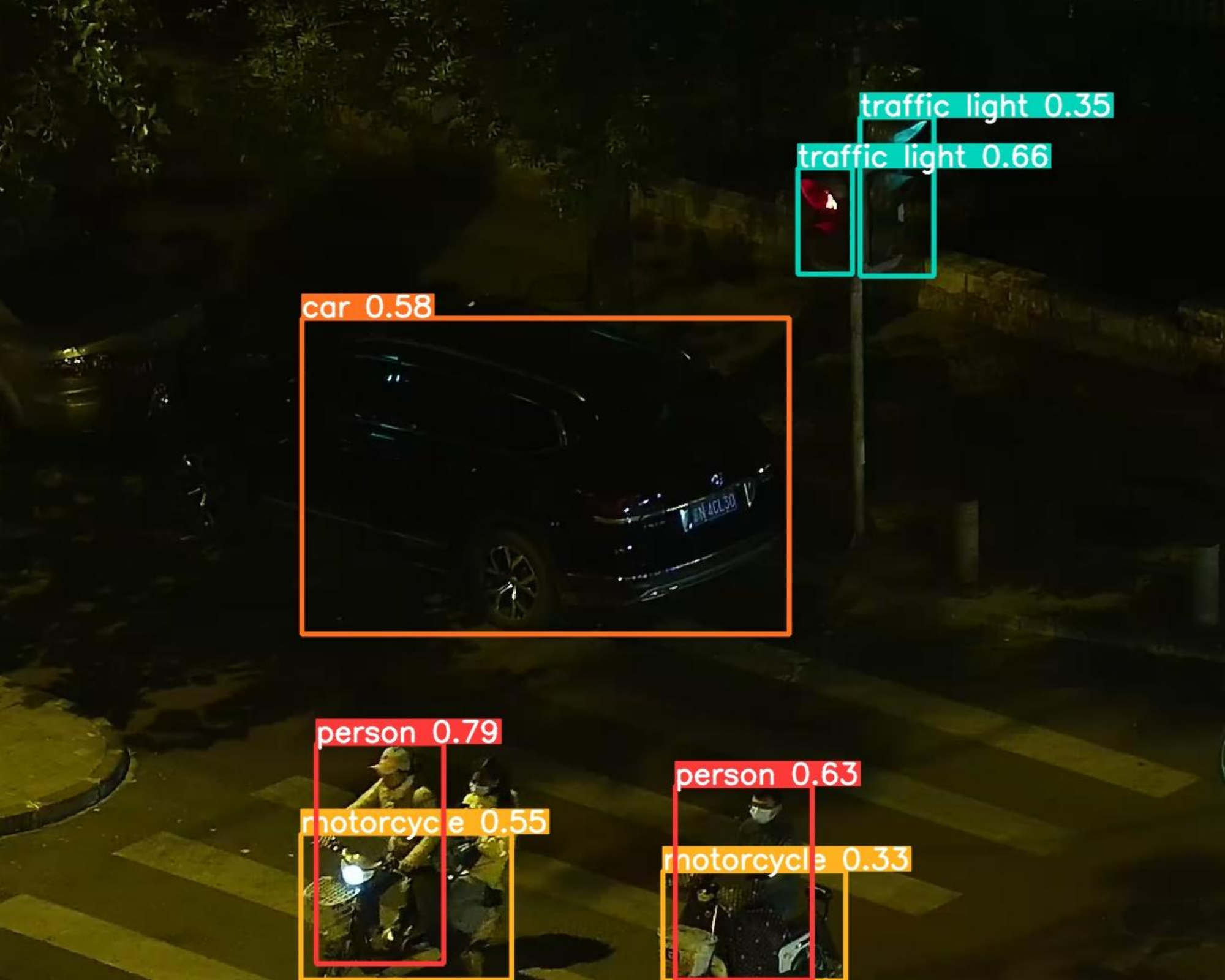}}
\subfloat[SeAFusion]{
		\includegraphics[width=1.1in]{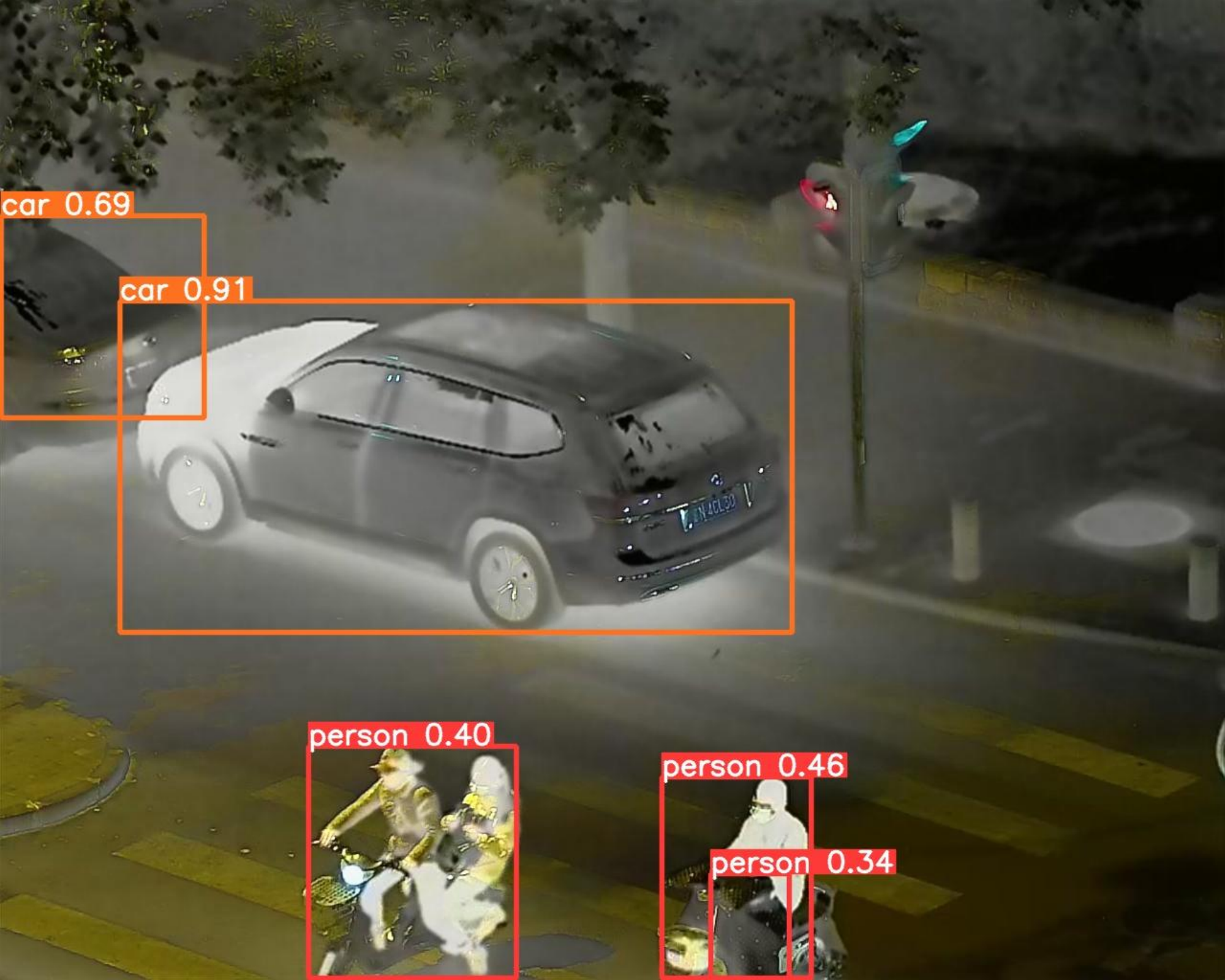}}
\subfloat[SCI+DenseFuse]{
		\includegraphics[width=1.1in]{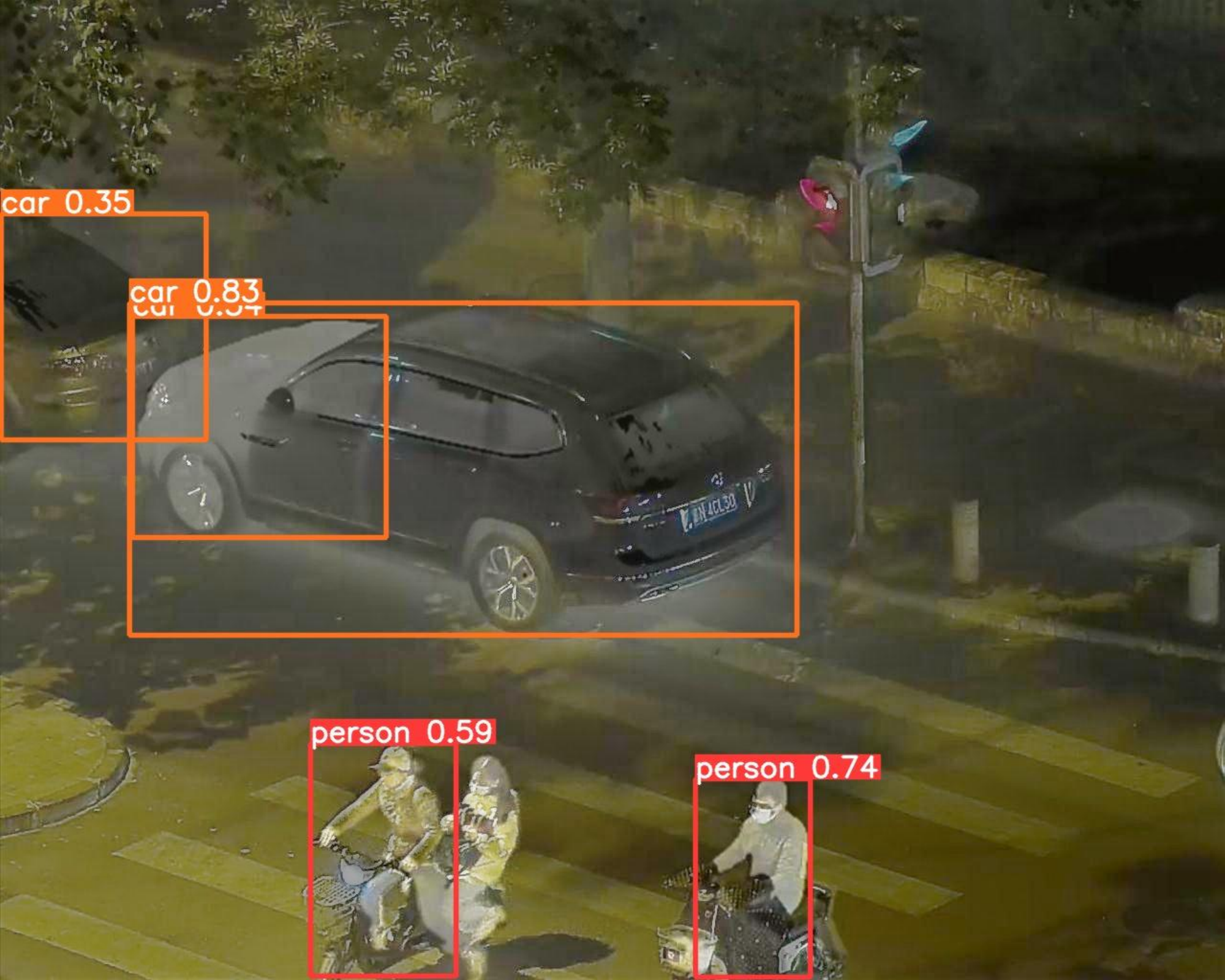}}
\\
\vspace{-2mm}
\subfloat[Zero-DCE+TarDAL]{
		\includegraphics[width=1.1in]{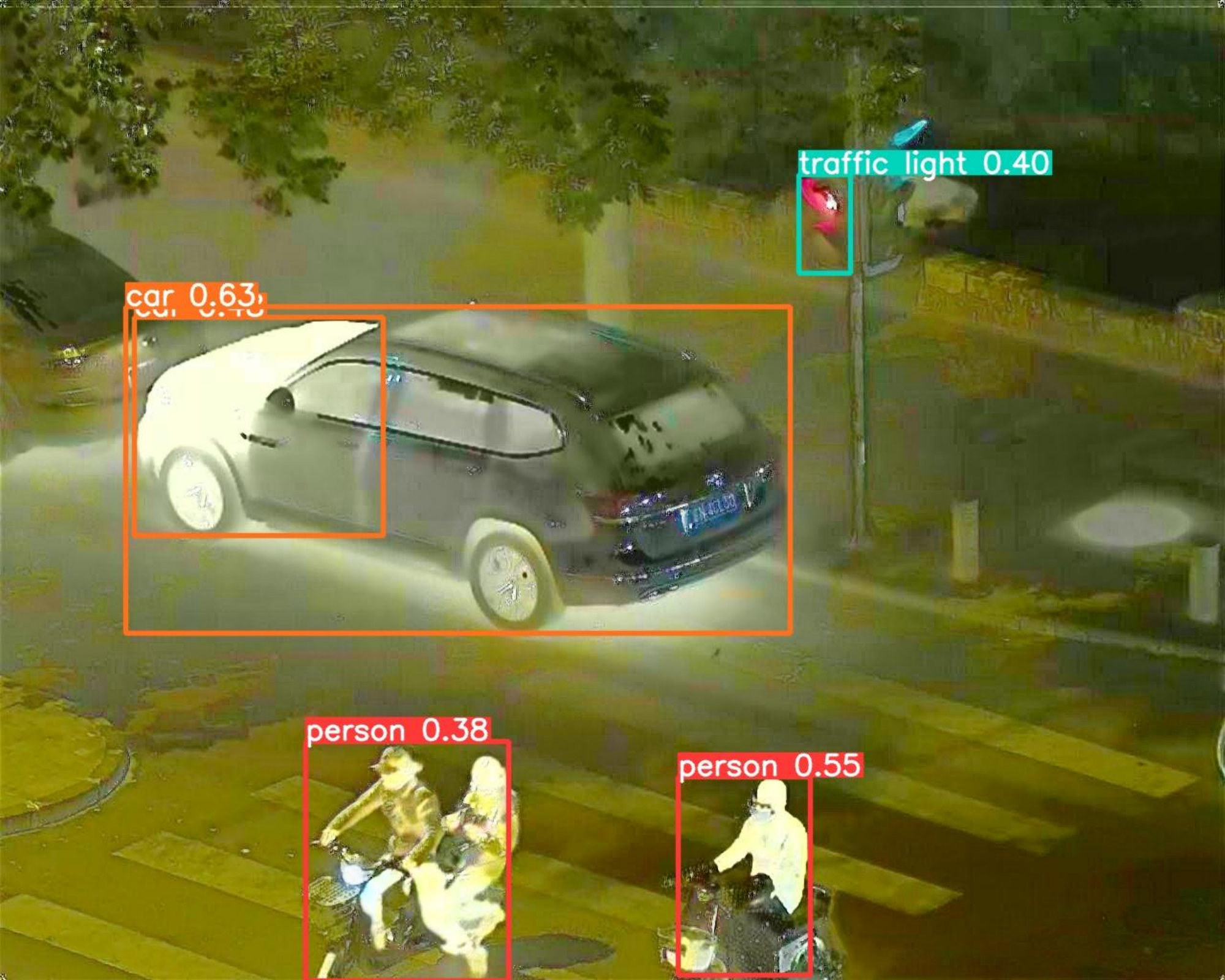}}
\subfloat[DIVFusion]{
		\includegraphics[width=1.1in]{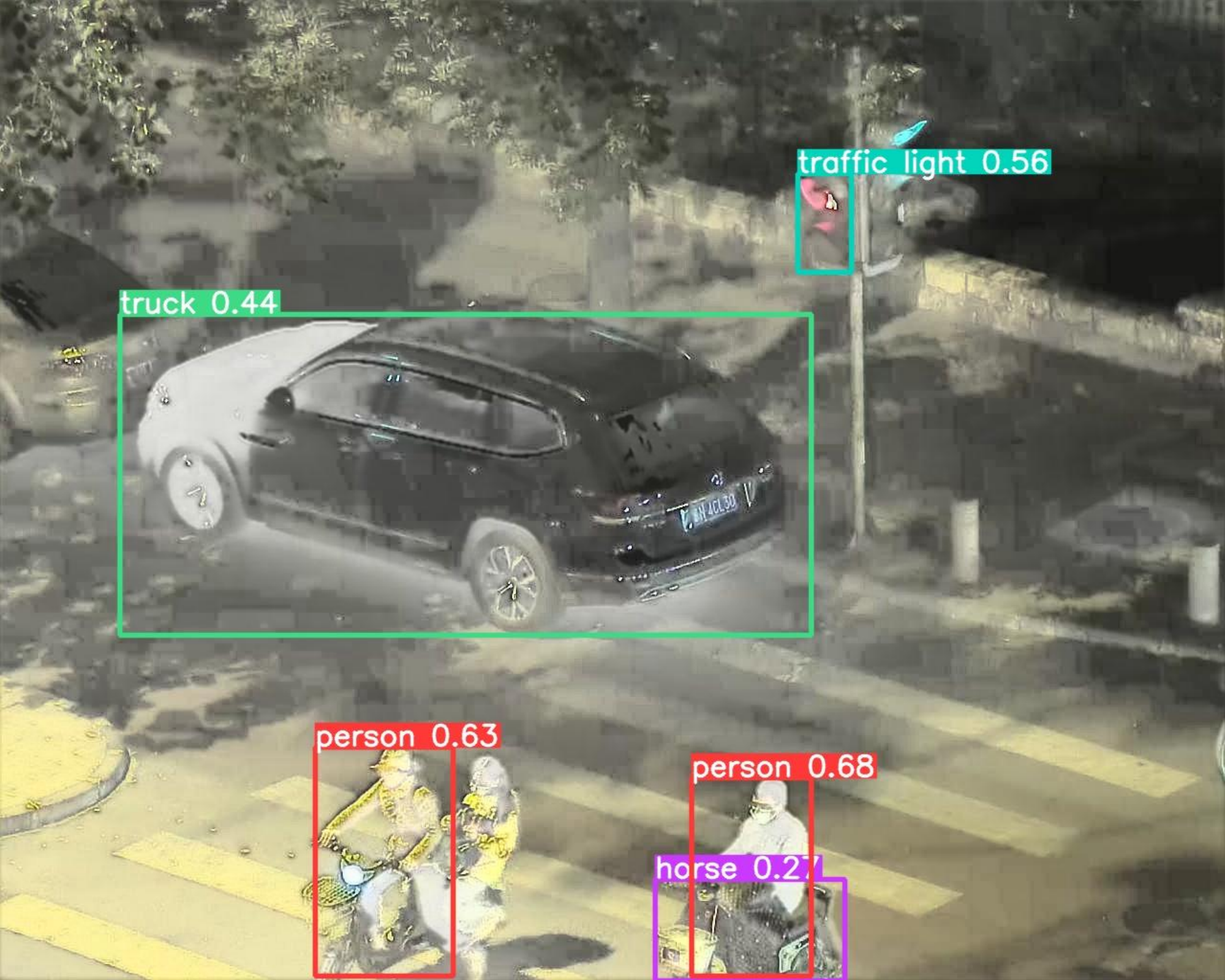}}
\subfloat[Ours]{
		\includegraphics[width=1.1in]{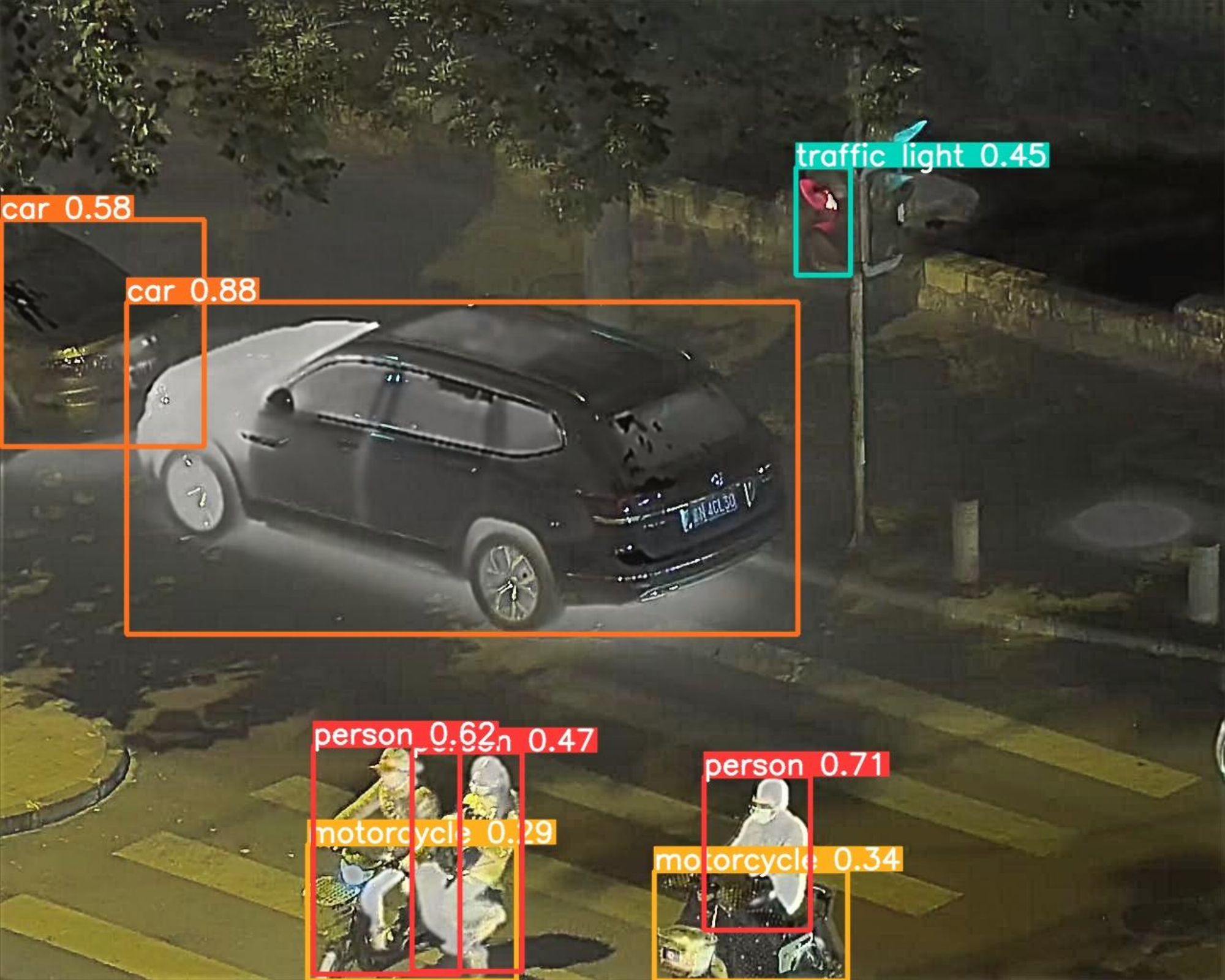}}
\vspace{5pt}
\caption{Detection performance of our fused images with four SOTA fusion results on the LLVIP dataset. \textbf{Our method can accurately predict motorcycles, with the maximum mAP of 0.88 for vehicle recognition.}}
\vspace{-15pt}
\label{fig:detect_2}
\end{figure}

\subsection{Performance Analysis on Detection Tasks}
To further validate the practicality of our method in high-level vision tasks, we apply the fused image to a typical computer vison task, \textit{i.e.}, object detection and pedestrian recognition. Concretely, we deploy the SOTA detection model YOLOv5 to conduct object detection and pedestrian recognition on nighttime visible images, and fused images. 

As illustrated in Fig. \ref{fig:detect_2}, the YOLOv5 detector is used to identify pedestrians, vehicles, and traffic signs on the street. In the detection results, the visible image cannot accentuate all objects effectively because of the illumination degradation. SeAfusion can compensate for the missing thermal imaging targets with infrared images, but it still misses some non-thermal imaging targets such as traffic lights. In contrast, our DFVO not only accurately identifies pedestrians in the scene but also precisely detects vehicles and traffic signals. For the pedestrian recognition task, we utilize quantitative experiments to assess the performance of fusion methods. In Table \ref{tab:table5}, our DFVO has the highest values in both Recall and Precision, indicating that our fused images can detect pedestrians more accurately with lower false positive rates. Furthermore, our fusion results also exhibit significant advantages in terms of mAP values, suggesting that the fused images generated by our method can provide better scenes for pedestrian recognition.

\begin{table}[!t]
\renewcommand{\arraystretch}{1.25}
\caption{
Pedestrian detection performance for visible, and fused images on the LLVIP dataset.
\label{tab:table5}}
\centering
\resizebox{\linewidth}{!}{
\begin{tabular}{cccccc}
\hline
Method        & Precision & Recall  & mAP@.5  & mAP@.7 & mAP@.5:.95  \\ \hline
VI & 0.849     & 0.682   & 0.746   & 0.525  & 0.372 \\
SeAFusion     & 0.889     & \uline{0.727}   & 0.784   & 0.516  & 0.460 \\
U2Fusion      & 0.885     & 0.697   & 0.809   & 0.528  & \uline{0.462} \\
SCI+DenseFuse & 0.880     & 0.667   & 0.758   & 0.533  & 0.416 \\
DIVFusion     & \uline{0.922} & 0.712   & \uline{0.817}   & \textbf{0.590}  & 0.429 \\
LENFusion     & 0.918     & 0.682       & 0.769           & \uline{0.554}           & 0.445 \\
Ours          & \textbf{0.961} & \textbf{0.758} & \textbf{0.861}   & 0.522  & \textbf{0.479} \\ \hline
\end{tabular}}
\vspace{-12pt}
\end{table}

\subsection{Discussion of Limitation}
On the one hand, the amount of parameters in our model is too large (the spatial complexity of CAM is $Q(n^2)$), placing the model at a disadvantage in terms of spatial complexity, which poses a significant challenge when applying it to the field of autonomous driving. On the other hand, some of the quantitative metrics in Table \ref{tab:table1} and Table \ref{tab:table2}, \textit{i.e.}, EN, SD and VIF are slightly lower compared to SOTA fusion methods. This is because our network prioritizes gradient information, texture details, and image distortion rate, but it focuses less on contrast and similarity information. In the future, we aim to further design an efficient lightweight feature extraction model to replace LCFE, with a greater emphasis on contrast and similarity information, aiming to further enhance the interactive information of the fused images. 

\section{Conclusion}
\label{sec:conclusion}
In this paper, we propose a novel holistic visible and infrared image fusion network, which achieves learning of illumination enhancement and image fusion simultaneously. Through qualitative and quantitative experiments comparing with SOTA fusion methods, we validate the advantages of our approach in visual scene perception and fused image clarity. Furthermore, the performance in object detection and pedestrian recognition tasks also demonstrates the potential of our method in high-level computer vision applications.

\bibliographystyle{IEEEtran}
\bibliography{IEEEabrv,Ref}

\begin{thebibliography}{10}
\providecommand{\url}[1]{#1}
\csname url@samestyle\endcsname
\providecommand{\newblock}{\relax}
\providecommand{\bibinfo}[2]{#2}
\providecommand{\BIBentrySTDinterwordspacing}{\spaceskip=0pt\relax}
\providecommand{\BIBentryALTinterwordstretchfactor}{4}
\providecommand{\BIBentryALTinterwordspacing}{\spaceskip=\fontdimen2\font plus
\BIBentryALTinterwordstretchfactor\fontdimen3\font minus
  \fontdimen4\font\relax}
\providecommand{\BIBforeignlanguage}[2]{{%
\expandafter\ifx\csname l@#1\endcsname\relax
\typeout{** WARNING: IEEEtran.bst: No hyphenation pattern has been}%
\typeout{** loaded for the language `#1'. Using the pattern for}%
\typeout{** the default language instead.}%
\else
\language=\csname l@#1\endcsname
\fi
#2}}
\providecommand{\BIBdecl}{\relax}
\BIBdecl

\bibitem{ma2021stdfusionnet}
J.~Ma, L.~Tang, M.~Xu, H.~Zhang, and G.~Xiao, ``Stdfusionnet: An infrared and
  visible image fusion network based on salient target detection,'' \emph{IEEE
  Transactions on Instrumentation and Measurement}, vol.~70, pp. 1--13, 2021.

\bibitem{shi2024diff}
Y.~Shi, Y.~Lin, P.~Wei, X.~Xian, T.~Chen, and L.~Lin, ``Diff-mosaic: augmenting
  realistic representations in infrared small target detection via diffusion
  prior,'' \emph{IEEE Transactions on Geoscience and Remote Sensing}, 2024.

\bibitem{lu2024sirst}
Y.~Lu, Y.~Lin, H.~Wu, X.~Xian, Y.~Shi, and L.~Lin, ``Sirst-5k: Exploring
  massive negatives synthesis with self-supervised learning for robust infrared
  small target detection,'' \emph{IEEE Transactions on Geoscience and Remote
  Sensing}, 2024.

\bibitem{liu2023yolactfusion}
C.~Liu, Q.~Feng, Y.~Sun, Y.~Li, M.~Ru, and L.~Xu, ``Yolactfusion: An instance
  segmentation method for rgb-nir multimodal image fusion based on an attention
  mechanism,'' \emph{Computers and Electronics in Agriculture}, vol. 213, p.
  108186, 2023.

\bibitem{bhatnagar2015novel}
G.~Bhatnagar and Z.~Liu, ``A novel image fusion framework for night-vision
  navigation and surveillance,'' \emph{Signal, Image and Video Processing},
  vol.~9, pp. 165--175, 2015.

\bibitem{da2006nonsubsampled}
A.~L. Da~Cunha, J.~Zhou, and M.~N. Do, ``The nonsubsampled contourlet
  transform: theory, design, and applications,'' \emph{IEEE transactions on
  image processing}, vol.~15, no.~10, pp. 3089--3101, 2006.

\bibitem{shi2025crossfuse}
Y.~Shi, C.~Shi, Z.~Weng, Y.~Tian, X.~Xian, and L.~Lin, ``Crossfuse: Learning
  infrared and visible image fusion by cross-sensor top-k vision alignment and
  beyond,'' \emph{IEEE Transactions on Circuits and Systems for Video
  Technology}, 2025.

\bibitem{han2013fast}
J.~Han, E.~J. Pauwels, and P.~De~Zeeuw, ``Fast saliency-aware multi-modality
  image fusion,'' \emph{Neurocomputing}, vol. 111, pp. 70--80, 2013.

\bibitem{fu2008multiple}
Y.~Fu, L.~Cao, G.~Guo, and T.~S. Huang, ``Multiple feature fusion by subspace
  learning,'' in \emph{Proceedings of the 2008 international conference on
  Content-based image and video retrieval}, 2008, pp. 127--134.

\bibitem{zhang2019infrared}
C.~Zhang, Z.~Yue, D.~Yan, and X.~Yang, ``Infrared and visible image fusion
  using joint convolution sparse coding,'' in \emph{2019 International
  Conference on Image and Video Processing, and Artificial Intelligence}, vol.
  11321.\hskip 1em plus 0.5em minus 0.4em\relax SPIE, 2019, pp. 181--189.

\bibitem{zang2021ufa}
Y.~Zang, D.~Zhou, C.~Wang, R.~Nie, and Y.~Guo, ``Ufa-fuse: A novel deep
  supervised and hybrid model for multifocus image fusion,'' \emph{IEEE
  Transactions on Instrumentation and Measurement}, vol.~70, pp. 1--17, 2021.

\bibitem{zhang2020ifcnn}
Y.~Zhang, Y.~Liu, P.~Sun, H.~Yan, X.~Zhao, and L.~Zhang, ``Ifcnn: A general
  image fusion framework based on convolutional neural network,''
  \emph{Information Fusion}, vol.~54, pp. 99--118, 2020.

\bibitem{ren2021infrared}
L.~Ren, Z.~Pan, J.~Cao, and J.~Liao, ``Infrared and visible image fusion based
  on variational auto-encoder and infrared feature compensation,''
  \emph{Infrared Physics \& Technology}, vol. 117, p. 103839, 2021.

\bibitem{li2023mrfddgan}
J.~Li, B.~Li, Y.~Jiang, L.~Tian, and W.~Cai, ``Mrfddgan: Multireceptive field
  feature transfer and dual discriminator-driven generative adversarial network
  for infrared and color visible image fusion,'' \emph{IEEE Transactions on
  Instrumentation and Measurement}, vol.~72, pp. 1--28, 2023.

\bibitem{li2022cgtf}
J.~Li, J.~Zhu, C.~Li, X.~Chen, and B.~Yang, ``Cgtf: Convolution-guided
  transformer for infrared and visible image fusion,'' \emph{IEEE Transactions
  on Instrumentation and Measurement}, vol.~71, pp. 1--14, 2022.

\bibitem{wang2022swinfuse}
Z.~Wang, Y.~Chen, W.~Shao, H.~Li, and L.~Zhang, ``Swinfuse: A residual swin
  transformer fusion network for infrared and visible images,'' \emph{IEEE
  Transactions on Instrumentation and Measurement}, vol.~71, pp. 1--12, 2022.

\bibitem{tang2023divfusion}
L.~Tang, X.~Xiang, H.~Zhang, M.~Gong, and J.~Ma, ``Divfusion: Darkness-free
  infrared and visible image fusion,'' \emph{Information Fusion}, vol.~91, pp.
  477--493, 2023.

\bibitem{zhang2024ev}
X.~Zhang, X.~Wang, C.~Yan, and Q.~Sun, ``Ev-fusion: A novel infrared and
  low-light color visible image fusion network integrating unsupervised visible
  image enhancement,'' \emph{IEEE Sensors Journal}, 2024.

\bibitem{gao2023l2fusion}
X.~Gao, G.~Lv, A.~Dong, Z.~Wei, and J.~Cheng, ``L2fusion: Low-light oriented
  infrared and visible image fusion,'' in \emph{2023 IEEE International
  Conference on Image Processing (ICIP)}.\hskip 1em plus 0.5em minus
  0.4em\relax IEEE, 2023, pp. 2405--2409.

\bibitem{chen2024lenfusion}
J.~Chen, L.~Yang, W.~Liu, X.~Tian, and J.~Ma, ``Lenfusion: A joint low-light
  enhancement and fusion network for nighttime infrared and visible image
  fusion,'' \emph{IEEE Transactions on Instrumentation and Measurement}, 2024.

\bibitem{wang2023enlighten}
H.~Wang, C.~Shu, and X.~Li, ``Enlighten fusion multiscale network for infrared
  and visible image fusion in dark environments,'' \emph{IEEE Signal Processing
  Letters}, 2023.

\bibitem{xian2024crose}
X.~Xian, Q.~Zhou, J.~Qin, X.~Yang, Y.~Tian, Y.~Shi, and D.~Tian, ``Crose:
  Low-light enhancement by cross-sensor interaction for nighttime driving
  scenes,'' \emph{Expert Systems with Applications}, vol. 248, p. 123470, 2024.

\bibitem{shi2024nitedr}
C.~Shi, L.~Fang, H.~Wu, X.~Xian, Y.~Shi, and L.~Lin, ``Nitedr: Nighttime image
  de-raining with cross-view sensor cooperative learning for dynamic driving
  scenes,'' \emph{IEEE Transactions on Multimedia}, 2024.

\bibitem{yang2023reference}
X.~Yang, J.~Gong, L.~Wu, Z.~Yang, Y.~Shi, and F.~Nie, ``Reference-free
  low-light image enhancement by associating hierarchical wavelet
  representations,'' \emph{Expert Systems with Applications}, vol. 213, p.
  118920, 2023.

\bibitem{carter2007introduction}
T.~Carter, ``An introduction to information theory and entropy,'' \emph{Complex
  systems summer school, Santa Fe}, 2007.

\bibitem{li2018densefuse}
H.~Li and X.-J. Wu, ``Densefuse: A fusion approach to infrared and visible
  images,'' \emph{IEEE Transactions on Image Processing}, vol.~28, no.~5, pp.
  2614--2623, 2018.

\bibitem{xu2020u2fusion}
H.~Xu, J.~Ma, J.~Jiang, X.~Guo, and H.~Ling, ``U2fusion: A unified unsupervised
  image fusion network,'' \emph{IEEE Transactions on Pattern Analysis and
  Machine Intelligence}, vol.~44, no.~1, pp. 502--518, 2020.

\bibitem{ma2020ganmcc}
J.~Ma, H.~Zhang, Z.~Shao, P.~Liang, and H.~Xu, ``Ganmcc: A generative
  adversarial network with multiclassification constraints for infrared and
  visible image fusion,'' \emph{IEEE Transactions on Instrumentation and
  Measurement}, vol.~70, pp. 1--14, 2020.

\bibitem{liu2022target}
J.~Liu, X.~Fan, Z.~Huang, G.~Wu, R.~Liu, W.~Zhong, and Z.~Luo, ``Target-aware
  dual adversarial learning and a multi-scenario multi-modality benchmark to
  fuse infrared and visible for object detection,'' in \emph{Proceedings of the
  IEEE/CVF conference on computer vision and pattern recognition}, 2022, pp.
  5802--5811.

\bibitem{tang2022image}
L.~Tang, J.~Yuan, and J.~Ma, ``Image fusion in the loop of high-level vision
  tasks: A semantic-aware real-time infrared and visible image fusion
  network,'' \emph{Information Fusion}, vol.~82, pp. 28--42, 2022.

\bibitem{dinh2014nice}
L.~Dinh, D.~Krueger, and Y.~Bengio, ``Nice: Non-linear independent components
  estimation,'' \emph{arXiv preprint arXiv:1410.8516}, 2014.

\bibitem{li2023iscmis}
F.~Li, Y.~Sheng, X.~Zhang, and C.~Qin, ``iscmis: Spatial-channel attention
  based deep invertible network for multi-image steganography,'' \emph{IEEE
  Transactions on Multimedia}, 2023.

\bibitem{xie2021enhanced}
Y.~Xie, K.~L. Cheng, and Q.~Chen, ``Enhanced invertible encoding for learned
  image compression,'' in \emph{Proceedings of the 29th ACM international
  conference on multimedia}, 2021, pp. 162--170.

\bibitem{xiao2020invertible}
M.~Xiao, S.~Zheng, C.~Liu, Y.~Wang, D.~He, G.~Ke, J.~Bian, Z.~Lin, and T.-Y.
  Liu, ``Invertible image rescaling,'' in \emph{Computer Vision--ECCV 2020:
  16th European Conference, Glasgow, UK, August 23--28, 2020, Proceedings, Part
  I 16}.\hskip 1em plus 0.5em minus 0.4em\relax Springer, 2020, pp. 126--144.

\bibitem{ardizzone2019guided}
L.~Ardizzone, C.~L{\"u}th, J.~Kruse, C.~Rother, and U.~K{\"o}the, ``Guided
  image generation with conditional invertible neural networks,'' \emph{arXiv
  preprint arXiv:1907.02392}, 2019.

\bibitem{zhuang2019invertible}
J.~Zhuang, N.~C. Dvornek, X.~Li, P.~Ventola, and J.~S. Duncan, ``Invertible
  network for classification and biomarker selection for asd,'' in
  \emph{Medical Image Computing and Computer Assisted Intervention--MICCAI
  2019: 22nd International Conference, Shenzhen, China, October 13--17, 2019,
  Proceedings, Part III 22}.\hskip 1em plus 0.5em minus 0.4em\relax Springer,
  2019, pp. 700--708.

\bibitem{vaswani2017attention}
A.~Vaswani, N.~Shazeer, N.~Parmar, J.~Uszkoreit, L.~Jones, A.~N. Gomez,
  {\L}.~Kaiser, and I.~Polosukhin, ``Attention is all you need,''
  \emph{Advances in neural information processing systems}, vol.~30, 2017.

\bibitem{li2020multigrained}
J.~Li, H.~Huo, C.~Li, R.~Wang, C.~Sui, and Z.~Liu, ``Multigrained attention
  network for infrared and visible image fusion,'' \emph{IEEE Transactions on
  Instrumentation and Measurement}, vol.~70, pp. 1--12, 2020.

\bibitem{chen2021defect}
R.~Chen, D.~Cai, X.~Hu, Z.~Zhan, and S.~Wang, ``Defect detection method of
  aluminum profile surface using deep self-attention mechanism under hybrid
  noise conditions,'' \emph{IEEE Transactions on Instrumentation and
  Measurement}, vol.~70, pp. 1--9, 2021.

\bibitem{dosovitskiy2020image}
A.~Dosovitskiy, L.~Beyer, A.~Kolesnikov, D.~Weissenborn, X.~Zhai,
  T.~Unterthiner, M.~Dehghani, M.~Minderer, G.~Heigold, S.~Gelly \emph{et~al.},
  ``An image is worth 16x16 words: Transformers for image recognition at
  scale,'' \emph{arXiv preprint arXiv:2010.11929}, 2020.

\bibitem{zhang2024perceive}
X.~Zhang, J.~Ma, G.~Wang, Q.~Zhang, H.~Zhang, and L.~Zhang, ``Perceive-ir:
  Learning to perceive degradation better for all-in-one image restoration,''
  \emph{arXiv preprint arXiv:2408.15994}, 2024.

\bibitem{zhang2022transformer}
J.~Zhang, A.~Liu, D.~Wang, Y.~Liu, Z.~J. Wang, and X.~Chen, ``Transformer-based
  end-to-end anatomical and functional image fusion,'' \emph{IEEE Transactions
  on Instrumentation and Measurement}, vol.~71, pp. 1--11, 2022.

\bibitem{liu2023transcending}
Y.~Liu, Z.~Xiong, Y.~Yuan, and Q.~Wang, ``Transcending pixels: boosting
  saliency detection via scene understanding from aerial imagery,'' \emph{IEEE
  Transactions on Geoscience and Remote Sensing}, vol.~61, pp. 1--16, 2023.

\bibitem{tian2023vision}
Y.~Tian, H.~Meng, F.~Yuan, Y.~Ling, and N.~Yuan, ``Vision transformer with
  enhanced self-attention for few shot ship target recognition in complex
  environments,'' \emph{IEEE Transactions on Instrumentation and Measurement},
  2023.

\bibitem{land1971lightness}
E.~H. Land and J.~J. McCann, ``Lightness and retinex theory,'' \emph{Josa},
  vol.~61, no.~1, pp. 1--11, 1971.

\bibitem{sandler2018mobilenetv2}
M.~Sandler, A.~Howard, M.~Zhu, A.~Zhmoginov, and L.-C. Chen, ``Mobilenetv2:
  Inverted residuals and linear bottlenecks,'' in \emph{Proceedings of the IEEE
  conference on computer vision and pattern recognition}, 2018, pp. 4510--4520.

\bibitem{zhang2019kindling}
Y.~Zhang, J.~Zhang, and X.~Guo, ``Kindling the darkness: A practical low-light
  image enhancer,'' in \emph{Proceedings of the 27th ACM international
  conference on multimedia}, 2019, pp. 1632--1640.

\bibitem{wei2018deep}
C.~Wei, W.~Wang, W.~Yang, and J.~Liu, ``Deep retinex decomposition for
  low-light enhancement,'' \emph{arXiv preprint arXiv:1808.04560}, 2018.

\bibitem{zhang2023self}
F.~Zhang, Y.~Shao, Y.~Sun, C.~Gao, and N.~Sang, ``Self-supervised low-light
  image enhancement via histogram equalization prior,'' in \emph{Chinese
  Conference on Pattern Recognition and Computer Vision (PRCV)}.\hskip 1em plus
  0.5em minus 0.4em\relax Springer, 2023, pp. 63--75.

\bibitem{he2016deep}
K.~He, X.~Zhang, S.~Ren, and J.~Sun, ``Deep residual learning for image
  recognition,'' in \emph{Proceedings of the IEEE conference on computer vision
  and pattern recognition}, 2016, pp. 770--778.

\bibitem{he2010single}
K.~He, J.~Sun, and X.~Tang, ``Single image haze removal using dark channel
  prior,'' \emph{IEEE transactions on pattern analysis and machine
  intelligence}, vol.~33, no.~12, pp. 2341--2353, 2010.

\bibitem{jia2021llvip}
X.~Jia, C.~Zhu, M.~Li, W.~Tang, and W.~Zhou, ``Llvip: A visible-infrared paired
  dataset for low-light vision,'' in \emph{Proceedings of the IEEE/CVF
  international conference on computer vision}, 2021, pp. 3496--3504.

\bibitem{guo2020zero}
C.~Guo, C.~Li, J.~Guo, C.~C. Loy, J.~Hou, S.~Kwong, and R.~Cong,
  ``Zero-reference deep curve estimation for low-light image enhancement,'' in
  \emph{Proceedings of the IEEE/CVF conference on computer vision and pattern
  recognition}, 2020, pp. 1780--1789.

\bibitem{ma2022toward}
L.~Ma, T.~Ma, R.~Liu, X.~Fan, and Z.~Luo, ``Toward fast, flexible, and robust
  low-light image enhancement,'' in \emph{Proceedings of the IEEE/CVF
  conference on computer vision and pattern recognition}, 2022, pp. 5637--5646.

\end{thebibliography}

\end{document}